\documentclass[twoside,twocolumn]{article}

\usepackage{tabulary,graphicx,times,caption,fancyhdr,amsfonts,amssymb,amsbsy,latexsym,amsmath}
\usepackage[utf8]{inputenc}
\usepackage{url,multirow,morefloats,floatflt,cancel,tfrupee,textcomp,colortbl,xcolor,pifont}
\usepackage[nointegrals]{wasysym}

\def\rvx{{\mathbf{x}}}
\newcommand{\D}{\mathcal{D}}
\newcommand{\E}{\mathbb{E}}
\newcommand{\Ls}{\mathcal{L}}
\newcommand{\KL}{\D_{\mathrm{KL}}}

\newcommand{\Tabref}[1]{Table \ref{#1}}

\def\eqref#1{equation~\ref{#1}}
\def\Eqref#1{Equation~\ref{#1}}

\def\Secref#1{Section~\ref{#1}}

\newcommand{\ie}{{\em i.e.}}
\newcommand{\eg}{{\em e.g.}}

\newcommand{\etc}{{\em etc.}}

\def\vtheta{{\bm{\theta}}}
\def\vphi{{\bm \phi}}

\newcommand{\argmax}{\operatornamewithlimits{argmax}}

\usepackage{multirow}
\usepackage{extarrows}
\usepackage{booktabs}
\usepackage{comment}
\usepackage{bm}

\usepackage{lineno,hyperref}
\modulolinenumbers[5]

\makeatletter

\usepackage{ifxetex}
\ifxetex\else
  \usepackage{dblfloatfix}
\fi

\@ifundefined{subparagraph}{
\def\subparagraph{\@startsection{paragraph}{5}{2\parindent}{0ex plus 0.1ex minus 0.1ex}%
{0ex}{\normalfont\small\itshape}}%
}{}

\def\URL#1#2{\@ifundefined{href}{#2}{\href{#1}{#2}}}

\def\UrlOrds{\do\*\do\-\do\~\do\'\do\"\do\-}%
\g@addto@macro{\UrlBreaks}{\UrlOrds}

\makeatother

\usepackage[paperheight=11in,paperwidth=8.3in,margin=2.5cm,headsep=.7cm,top=2.5cm]{geometry}
\usepackage[T1]{fontenc}

\renewenvironment{abstract}
	{\trivlist\item[]\leftskip0pt\par\vskip4pt\noindent
  	\textbf{\abstractname}\mbox{\null}\\}
	{\par\noindent\endtrivlist}

\def\keywords#1{\par\medskip\par\noindent\textbf{Keywords}: #1\par}

 \date{} \emergencystretch 8pt

\makeatletter
\def\author#1{\gdef\@author{\hskip-\tabcolsep%
	\parbox{\textwidth}{\raggedright\bfseries#1\\[1pc]}}}
\def\address[#1]#2{\g@addto@macro\@author{\\\hskip-\tabcolsep\parbox{\textwidth}{\raggedright%
	\normalsize\normalfont\textsuperscript{#1}#2}}}
\let\addresslink\textsuperscript
\def\correspondence#1{\g@addto@macro\@author{\\\hskip-\tabcolsep\parbox{\textwidth}{\raggedright%
	\vspace*{10pt}\normalsize\normalfont~\\#1~\\[12pt]}}}
\def\email#1{\g@addto@macro\@author{\\\hskip-\tabcolsep\parbox{\textwidth}{\raggedright%
	\normalsize\normalfont Emails: #1}}}

\def\title#1{\gdef\@title{\vspace*{-30pt}%
	\raggedright\textbf{\@journaltitle}~\\%
  \raggedright\bfseries\ifx\@articleType\@empty\vspace*{20pt}\else%
  \vspace*{20pt}\@articleType\vspace*{20pt}\\\fi#1}}
\let\@journaltitle\@empty \def\journaltitle#1{\gdef\@journaltitle{{\normalfont\itshape#1}}}
\let\@articleType\@empty \def\articletype#1{\gdef\@articleType{{\normalfont\itshape#1}}}

\let\@runningHead\@empty \def\RunningHead#1{\gdef\@runningHead{{\normalfont #1}}}

\usepackage{fancyhdr}
\fancypagestyle{headings}{\fancyhf{}
  \fancyhead[R]{\itshape\@runningHead}
  \fancyfoot[C]{\thepage}}
\usepackage[%
	numbers,sort&compress%
  ]{natbib}

\usepackage{float,xcolor}

\articletype{Research Article} 

\begin{document}

\title{Adversarial and Random Transformations for Robust Domain Adaptation and Generalization}

\author{%
	Liang Xiao\addresslink{1},
  	Jiaolong Xu\addresslink{1},
  	Dawei Zhao\addresslink{1},
  	Erke Shang\addresslink{1},
  	Qi Zhu\addresslink{1}
  	and
  	Bin Dai\addresslink{1}
    }
		
\address[1]{Unmanned Systems Research Center, National Innovation Institute of Defense Technology, Beijing 10071, China}

\email{xiaoliang.cs@gmail.com(Liang Xiao), jiaolong\_xu@126.com (Jiaolong Xu) }%


\maketitle 

\begin{abstract}
    Data augmentation has been widely used to improve generalization in training deep neural networks. Recent works show that using worst-case transformations or adversarial augmentation strategies can significantly improve the accuracy and robustness. However, due to the non-differentiable properties of image transformations, searching algorithms such as reinforcement learning or evolution strategy have to be applied, which are not computationally practical for large scale problems. In this work, we show that by simply applying consistency training with random data augmentation, state-of-the-art results on domain adaptation (DA) and generalization (DG) can be obtained.  To further improve the accuracy and robustness with adversarial examples, we propose a differentiable adversarial data augmentation method based on spatial transformer networks (STN). The combined adversarial and random transformations based method outperforms the state-of-the-art on multiple DA and DG benchmark datasets.
	Besides, the proposed method shows desirable robustness to corruption, which is also validated on commonly used datasets.

\end{abstract}

\section{Introduction}
For modern computer vision applications, we expect a  model trained on large scale datasets can perform uniformly well across various testing scenarios. For example, consider the perception system of a self-driving car, we wish it could generalize well across weather conditions and city environments. However, the current supervised learning based models remain weak at out-of-distribution generalization \cite{DAN:2015}.  When testing and training data are drawn from different distributions, the model can suffer from a significant accuracy drop. It is known as domain shift problem, which is drawing increasing attention in recent years \cite{DAN:2015} \cite{DANN:2016} \cite{ADDA:2017} \cite{cycada:2018} \cite{rot:2019}.

Domain adaptation (DA) and domain generalization (DG) are two typical techniques to address the domain shift problem. DA and DG aim to utilize one or multiple labeled source domains to learn a model that performs well on an unlabeled target domain. The major difference between DA and DG is that DA methods require target data during training whereas DG methods do not require target data in the training phase. DA can be categorized as supervised, semi-supervised and unsupervised, depending on the availability of the labels of target data. In this paper, we consider unsupervised DA which do no require label of target data. In recent years, many works have been proposed to address either DA or DG problem \cite{cycada:2018} \cite{carlucci:2019}. In this work, we address both DA and DG in a unified framework.

Data augmentation is an effective technique for reducing overfitting and has been widely used in many computer vision tasks to improve the generalization ability of the model. Recent studies show that using worst-case transformations or adversarial augmentation strategies can greatly improve the generalization and robustness of the model \cite{zhang:2019} \cite{Volpi_iccv:2019}. However, due to the non-differentiable properties of image transformations, searching algorithms such as reinforcement learning \cite{AutoAugment:2019} \cite{fast_auto_aug:2019} or evolution strategy \cite{Volpi_iccv:2019} have to be applied, which are not computationally practical for large scale problems. In this work, we are concerned with the effectiveness of data augmentation for DA and DG, especially the adversarial data augmentation strategies without using heavy searching based methods.

Motivated by the recent success of RandomAugment \cite{randaugment:2019} on improving the generalization of deep learning models and consistency training in semi-supervised and unsupervised learning \cite{Sajjadi:2016} \cite{uda:2019} \cite{Suzuki:2020}, we propose a unified DA and DG method by incorporating consistency training with random data augmentation. The idea is simple, when doing a forward pass in the neural network, we force the randomly augmented and non-augmented pair of training examples to have similar responses by applying a consistency loss. Because the consistency training does not require labeled examples, we can apply it with unlabeled target domain data for domain adaptation training. The consistency training and source domain supervised training are in a joint multi-task training framework and can be trained end-to-end. The random augmentation can also be regarded as a way of noisy injection and by applying consistency training with the noisy and original examples, the model's generalization ability is expected to be improved. Following VAT \cite{vat:2018} and UDA \cite{uda:2019}, we use the KL divergence to compute the consistency loss.

To further improve the accuracy and robustness, we consider employing adversarial augmentations to find worst-case transformations. Our interest is in performing adversarial augmentation for DA and/or DG without using the searching based methods. Most of the image transformations are non-differentiable, except a subset of geometric transformations. Inspired by the spatial transformer networks (STN) of \cite{stn:2015}, we propose a differentiable adversarial spatial transformer network for both DA and DG. As we will show in the experimental section, the adversarial STN alone achieves promising results on both DA and DG tasks. When combined with random image transformations, it outperforms state-of-the-art, which is validated on several DA and DG benchmark datasets.

In this work, apart from the cross-domain generalization ability, robustness is also our concern. This is particularly important for real applications when applying a model to unseen domains, which, however, is largely ignored in current DA and DG literature. We evaluate the robustness of our models on \textbf{CIFAR-10-C} \cite{cifar10-c},  which is a robustness benchmark with $15$ types of corruptions algorithmically simulated to mimic real-world corruptions. The experimental results show that our proposed method not only reduces the cross-domain accuracy drop but also improves the robustness of the model.

Our contributions can be summarized as follows:

\begin{itemize}

\item We build a unified framework for domain adaptation and domain generalization based on data augmentation and consistency training.
\item We propose an end-to-end differentiable adversarial data augmentation strategy with spatial transformer networks to improve the accuracy and robustness.
\item We show that our proposed methods outperform state-of-the-art DA and DG methods on multiple object recognition datasets.
\item We show that our model is robust to common corruptions and obtained promising results on \textbf{CIFAR-10-C} robustness benchmark.
\end{itemize}

\section{Related work}
\subsection{Domain adaptation}
Modern domain adaption methods usually address domain shift by learning domain invariant features. This purpose can be achieved by minimizing a certain measure of domain variance, such as the Maximum Mean Discrepancy (MMD)  \cite{DAN:2015} \cite{TzengHZSD14} and mutual information \cite{cinWangKe}, or aligning the second-order statistics of source and target distributions \cite{CORAL2016} \cite{DeepCORAL2016}.

Another line of work uses adversarial learning to learn features that are discriminative in source space and at the same time invariant with respect to domain shift \cite{Gani:2015} \cite{ZhangTMech} \cite{ZhangTII}. In \cite{DANN:2016}  and \cite{Gani:2015}, a gradient reverse layer is proposed to achieve domain adversarial learning by back-propagation. 
In \cite{FANG2021298}, a multi-layer adversarial DA method was proposed, in which a feature-level domain classifier is used to learn domain-invariant representation while a  prediction-level domain classifier is used to reduce domain discrepancy in the decision layer.
In \cite{cycada:2018}, CycleGAN \cite{CycleGAN} based unpaired image translation is employed to achieve both feature-level and pixel-level adaptation.
In \cite{DIRT-T:2018}, cluster assumption is applied to domain adaptation and a method called Virtual Adversarial Domain Adaptation (VADA) is proposed. VADA utilizes VAT \cite{vat:2018} to enforce classifier consistency within the vicinity of samples. Drop to Adapt \cite{DropAdapt:2019} also enforces cluster assumption by leveraging adversarial dropout.
In \cite{ALDA:aaai20}, adversarial learning and self-training are combined, in which an adversarial-learned confusion matrix is utilized to correct the pseudo label and then align the feature distribution.

Recently, self-supervised learning based domain adaptation has been proposed \cite{rot:2019}. Self-supervised DA integrates a pretext learning task, such as image rotation prediction in the target domain with the main task in the source domain.  Self-supervised DA has shown capable of learning domain invariant feature representations \cite{rot:2019} \cite{ss-da-consistency:2019}.

\subsection{Domain generalization}
Similar to domain adaptation, existing work usually learns domain invariant feature by minimizing the discrepancy between the given multiple source domains, assuming that the source-domain invariant feature works well for the unknown target domain. Domain-Invariant Component Analysis (DICA) is proposed in \cite{muandet13ICML} to learn an invariant transformation by minimizing the dissimilarity across domains. 
In \cite{DGautoencoders}, a multi-domain reconstruction auto-encoder is proposed to learn domain-invariant feature. 

Adversarial learning has also been applied in DG. In \cite{Li_2018_CVPR}, MMD-based adversarial autoencoder (AAE) is proposed to align the distributions among different domains, and match the aligned distribution to an arbitrary prior distribution. In \cite{correlation:2019}, correlation alignment is combining with adversarial learning to minimizing the domain discrepancy. In \cite{ZHOU2021469}, optimal transport with Wasserstein distance is adopted in adversarial learning framework to align the marginal feature distribution over all the
source domains.

Some work utilizes low-rank constraint to achieve domain generalization capability, such as \cite{Low-Rank:2014}, \cite{LRExeSVMs} and \cite{Ding2017DeepDG}. Meta learning has recently been applied to domain generalization, including \cite{metareg:2018}, \cite{MLDG_AAA18} and \cite{Dou:2019}. In \cite{CHEN2022418}, a method integrated adversarial learning and meta-learning was proposed. 

In \cite{carlucci:2019} \cite{jigrot:pami21}, self-supervised DG is proposed by introducing a secondary task to solve a jigsaw puzzle and / or predict image rotation. This auxiliary task helps the network to learn the concepts of spatial correlation while acting as a regularizer for the main task. With this simple model, state-of-the-art domain generalization performance can be achieved.

\subsection{Data augmentation}
Data augmentation is a widely used trick in training deep neural networks. 
In visual learning, early data augmentation usually uses a composition of elementary image transformation, including translation, flipping, rotation, stretching, shearing and adding noise  \cite{Dosovitskiy:2014}. Recently, more complicated data augmentation approaches have been proposed, such as CutOut \cite{Cutout}, Mixup \cite{mixup:2018} and AugMix \cite{augmix:2020}. These methods are designed by human experts based on prior knowledge of the task together with trial and error. 
To automatically find the best data augmentation method for specific task, policy search based automated data augmentation approaches have been proposed, such as \emph{AutoAugment} \cite{AutoAugment:2019}, and \emph{Population based augmentation (PBA)} \cite{pba:2019}. The main drawback of these automated data augmentation approaches is the prohibitively high computational cost. Recently, \cite{zhang:2019} improves the computational efficiency of \emph{AutoAugment} by simultaneously optimizing target related object and augmentation policy search loss.

Another kind of data augmentation methods aims at finding the worst-case transformation and utilize them to improve the robustness of the learned model. In \cite{volpi:2018}, adversarial data augmentation is employed to generate adversarial examples which are appended during training to improve the generalization ability. In \cite{Volpi_iccv:2019}, it further proposed searching for worst-case image transformations by random search or evolution-based search.  Reinforcement learning is used in \cite{zhang:2019} to search for adversarial examples, in which \emph{RandAugment} and worst-case transformation are combined. 

Recently, consistency training with data augmentation has been used for improving semi-supervised training \cite{uda:2019} and generalization ability of supervised training \cite{Ait2019}.

Most of the related works focus either on domain adaptation or domain generalization, while in this work we consider designing a general model to address both of them. Domain adversarial training is a widely used technique for DA and DG, while our work does not follow this mainstream methodology but seeks resolution from the aspect of representation learning, {\eg} self-supervised learning \cite{rot:2019} and consistency learning \cite{ss-da-consistency:2019}. For representation learning, data augmentation also plays an important role for it can reduce model overfitting and improve the generalization ability. However, whether data augmentation can address cross-domain adaptation and generalization problems is still not well explored. In this work, we design a framework to incorporate data augmentation and consistency learning to address both domain adaptation and generalization problems.  

\section{The Proposed Approach}
In this section, we present the proposed method for domain adaptation and generalization in detail. 

\subsection{Problem Statement}

In the domain adaptation and generalization problem, we are given a source domain $\mathit{D}_s$ and target domain $\mathit{D}_t$ containing samples from two different distributions $P_{S}$ and $P_{T}$ respectively.  Denoting by $\{\rvx^s, \hat{y}^s\} \in \mathit{D}_s$ a labeled source domain sample, and $\{\rvx^t\} \in \mathit{D}_t$ a target domain sample without label, we have $\rvx^s \sim P_{S}$, $\rvx^t \sim P_{T}$, and $P_S \neq P_T$. When applying the model trained on the source domain to target domain, the distribution mismatch can lead to  significant performance drop. 

The task of unsupervised domain adaptation is to train a classification model $F: \rvx^s \rightarrow y^s$ which is able to classify $\rvx^t$ to the corresponding label $y^t$ given $\{\rvx^s, \hat{y}^s\}$ and $\{\rvx^t\}$ as training data. On the other hand, the task of domain generalization is to train a classification model $F: \rvx^s \rightarrow y^s$ which is able to classify $\rvx^t$ to the corresponding label $y^t$ given only $\{\rvx^s, \hat{y}^s\}$. The difference between these two tasks is whether $\{\rvx^t\}$ is involved or not during training. For both domain adaptation and generalization, we assume there are $n_s$ source domains where $n_s \geqslant 1$ and there is one single target domain.

Many works have been proposed to address either domain adaptation or domain generalization. In this work, we propose a unified framework to address both problems. In this follows, we first focus on domain adaptation and introduce the main idea and explain the details of the proposed method. Then, we show how this method can be adapted to domain generalization tasks as well.

\begin{table*}[h!]
	\centering
	\begin{tabular}{l|l|l|l}
		\hline
		& Name & Magnitude type & Magnitude range \\ \hline
		\multirow{5}{*}{\shortstack[l]{Geometric\\transformations}}
		& \texttt{ShearX} & continuous & [0, 0.3]\\
		& \texttt{ShearY} & continuous & [0, 0.3]\\
		& \texttt{TranslateX} & continuous & [0, 100] \\
		& \texttt{TranslateY} & continuous & [0, 100] \\
		& \texttt{Rotate} & continuous & [0, 30] \\
		& \texttt{Flip}   & none & none \\ \hline
		\multirow{9}{*}{\shortstack[l]{Color-based\\transformations}}
		& \texttt{Solarize} & discrete & [0, 255]\\
		& \texttt{Posterize} & discrete & [0, 4] \\
		& \texttt{Invert} & none & none\\
		& \texttt{Contrast} & continuous & [0.1, 1.9] \\
		& \texttt{Color} & continuous & [0.1, 1.9] \\
		& \texttt{Brightness} & continuous & [0.1, 1.9] \\
		& \texttt{Sharpness} & continuous & [0.1, 1.9] \\
		& \texttt{AutoContrast} & none & none \\
		& \texttt{Equalize}   & none & none \\ \hline
		\multirow{2}{*}{Other transformations}
		& \texttt{CutOut} & discrete & [0, 40] \\
		& \texttt{SamplePairing} & continuous & [0, 0.4] \\
		\hline
	\end{tabular}
	\vspace{5pt}
	\caption{Image transform operations. Some operations have discrete magnitude parameters, while others have no or continuous magnitude parameters.}
	\label{tab:operations}
\end{table*}

\subsection{Random image transformation with consistency training}
\label{subsec:random_transformations}

\begin{figure}[tb]
      \centering
      \includegraphics[width=\linewidth]{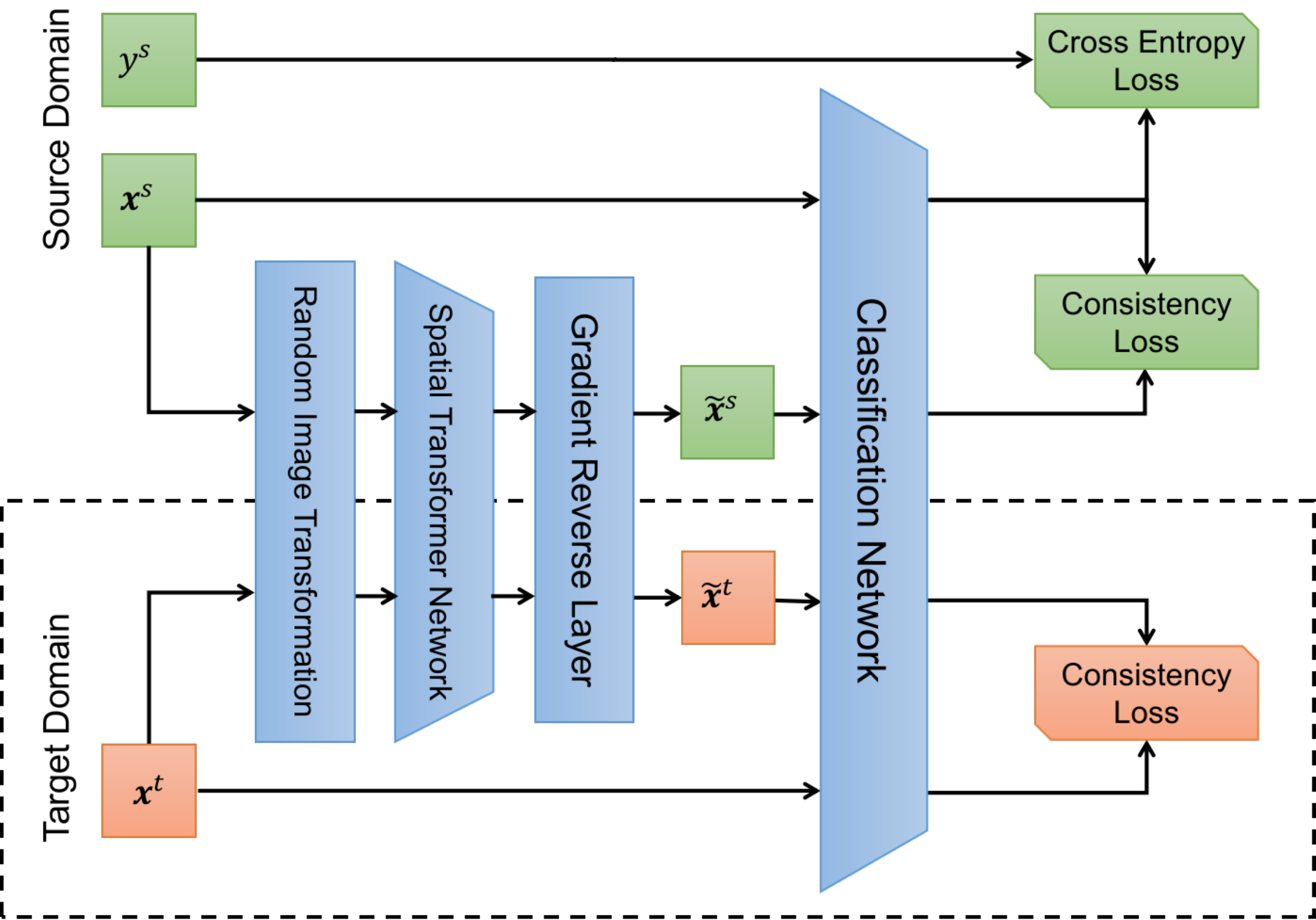}
      \caption{Overview of our proposed model. We propose to use random image transformations and adversarial spatial transformer networks (STN) to achieve domain adaptation and generalization (without the dash line bounding box).}
      \label{fig:uda}
\end{figure}

Inspired by a recent line of work \cite{uda:2019} \cite{ss-da-consistency:2019} in semi-supervised learning which incorporates consistency training with unlabeled examples to enforce the smoothness of the model, we propose to use image transformation as a way of noisy injection and apply consistency training with the noisy and original examples. The overview of the proposed random image transformation with consistency training for domain adaptation is depicted in \figurename~\ref{fig:uda}. In this section, we focus on the random image transformation part and leave the adversarial spatial transformer networks in the next section. The main idea can be explained as follows:

(1) Given an input image $\rvx$ either from source or target domain, we compute the output distribution $p(y \mid \rvx)$ with $\rvx$ and a noisy version $p(y \mid \tilde{\rvx})$ by applying random image transformation to $\rvx$;

(2) For domain adaptation, we jointly minimize the classification loss with labeled source domain samples and a divergence metric between the two distributions $\D(p(y \mid \rvx) \parallel  p(y \mid \tilde{\rvx}))$ with unlabeled source and target domain samples, where $\mathcal D$ is a discrepancy measure between two distributions;

(3) For domain generalization, the procedure is similar to (2) but without using any target domain samples.

Our intuition is that, on one hand, minimizing the consistency loss can enforce the model to be insensitive to the noise and improve the generalization ability; On the other hand, the consistency training gradually transmits label information from labeled source domain examples to unlabeled target domain ones, which improves the domain adaptation ability.



The applied random image transformations are similar to {\em RandAugment} \cite{randaugment:2019}. \Tabref{tab:operations} shows the types of image transformations used in this work. The images transformations are categorized into three groups. The first group is the geometric transformations, including \texttt{Shear}, \texttt{Translation}, \texttt{Rotation} and \texttt{Flip}. The second group is the color enhancing based image transformations, {\eg} \texttt{Solarize}, \texttt{Contrast} \etc, and the last group including other transformations {\eg} \texttt{CutOut} and \texttt{SamplePairing}. Each type of image transformation has a corresponding magnitude, which indicates the strength of the transformation. The magnitude can be either a continues or discrete variable. Following \cite{randaugment:2019}, we also normalize the magnitude to a range from $0$ to $10$, in order to employ a linear scale of magnitude for each type of transformations. In other words, a value of $10$ indicates the maximum scale for a given transformation, while $0$ means minimum scale. Note that, these image transformations are commonly used as searching policies in recent auto-augmentation literatures, such as \cite{AutoAugment:2019}, \cite{zhang:2019} and \cite{fast_auto_aug:2019}. Following \cite{randaugment:2019}, we do not use search, but instead uniformly sample from the same set of image transformations. Specifically, for each training sample, we uniformly sample $N_{aug}$ image transformations from \Tabref{tab:operations} with normalized magnitude value of $M_{aug}$ and then apply them to the image sequentially. $N_{aug}$ and $M_{aug}$ are hyper-parameters. Following the practice of \cite{randaugment:2019}, we sampled $N_{aug} \in \{1, 2, 3, 5, 10\}$ and $M_{aug} \in \{3, 6, 9, 12\}$. We conduct validation experiments on PACS dataset and VisDA dataset and find $N_{aug} = 2$ and $M_{aug} = 9$ obtains the best results, thus we keep  $N_{aug} = 2$ and $M_{aug} = 9$ in all our experiments.

Following VAT \cite{vat:2018} and UDA \cite{uda:2019}, we also use the KL divergence to compute the consistency loss. We denote by $\vtheta_{m}$ the parameters of the classification model. The classification loss with labeled source domain samples is written as following cross entropy loss:

\begin{equation}
      \label{eq:cross_entropy_loss}
      \mathcal{L}_{m}(\vtheta_{m}) = \E_{\rvx^s, \hat{y}^s \in \mathit{D}_s}[-\log{p(\hat{y}^s \mid \rvx^s)}].
\end{equation}

The consistency loss for domain adaptation can be written by:

\begin{equation}
\small
\label{eq:consistency_loss_da}
\Ls_{c}(\vtheta_{m}) = \E_{\rvx \in \mathit{D}_s \cup \mathit{D}_t} \E_{\tilde{\rvx} \in \tilde{\mathit{D}_s} \cup \tilde{\mathit{D}_t}} [\KL(\hat{p}(y \mid \rvx), p(y \mid \tilde{\rvx}))],
\end{equation}
where $\hat{p}(y \mid \rvx)$ uses a fixed copy of $\vtheta_{m}$ which means that the gradient is not propagated through $\hat{p}(y \mid \rvx)$. 

As a common underlying assumption in many semi-supervised learning methods, the classifier’s decision boundary should not pass through high-density regions of the marginal data distribution \cite{mixmatch:2019}. The conditional entropy minimization loss (EntMin) \cite{EntMin:2005} enforces this by encouraging the classifier to output low-entropy predictions on unlabeled data. EntMin is also combined with VAT in \cite{vat:2018} to obtain stronger results. Specifically, the conditional entropy minimization loss is written as:

\begin{equation}
\label{eq:loss_ent_min}
\mathcal{L}_{e}(\vtheta_{m}) = \mathbb{E}_{\rvx^t \in \mathit{D}_t }[-p(y^t \mid \rvx^t)\log{p(y^t \mid \rvx^t)}].
\end{equation}

Following \cite{carlucci:2019} and \cite{rot:2019}, we also apply the conditional entropy minimization loss to the unlabeled target domain data to minimize the classifier prediction uncertainty. The full objective of domain adaptation is thus written as follows:

\begin{equation}
      \label{eq:loss_da}
      \mathcal{J}_{DA}(\vtheta_{m}) = \min_{\vtheta_{m}} (\mathcal{L}_{m} + \lambda_c \mathcal{L}_{c} + \lambda_e \mathcal{L}_{e}),
\end{equation}
where $\lambda_c$ and $\lambda_e$ are the weight factor for the consistency loss and conditional entropy minimization loss. 

For domain generalization, as no target domain data is involved during the training, \Eqref{eq:consistency_loss_da} can be written as:

\begin{equation}
\small
      \label{eq:consistency_loss_dg}
      \Ls_{c}(\vtheta_{m}) = \E_{\rvx \in \mathit{D}_s} \E_{\tilde{\rvx} \in \tilde{\mathit{D}_s}} [\KL(\hat{p}(y \mid \rvx), p(y \mid \tilde{\rvx}^t))],
\end{equation}
and the final objective function is the weighted sum of \Eqref{eq:consistency_loss_dg} and the classification loss \Eqref{eq:cross_entropy_loss} :

\begin{equation}
      \label{eq:loss_dg}
      \mathcal{J}_{DG}(\vtheta_{m}) = \min_{\vtheta_{m}} (\mathcal{L}_{m} + \lambda_c \mathcal{L}_{c}).
\end{equation}

\subsection{Adversarial Spatial Transformer Networks}

The proposed random image transformation with consistency training is a simple and effective method to reduce domain shift. Recent works show that using worst-case transformations or adversarial augmentation strategies can significantly improve the accuracy and robustness of the model \cite{zhang:2019} \cite{Volpi_iccv:2019}. However, most of the image transformations in \Secref{subsec:random_transformations}
are non-differentiable, making it difficult to apply gradient descent based method to obtain optimal transformations. To address this problem, searching algorithms such as reinforcement learning \cite{zhang:2019} or evolution strategy \cite{Volpi_iccv:2019} are employed in recent works, which however are computationally expensive and don't guarantee to obtain the global optima. In this work, we find that a subset of the image transformations in \Tabref{tab:operations} are actually differentiable, {\ie}, the geometric transformations. In this work, we build our adversarial geometric transformation on top of the spatial transformer networks (STN) \cite{stn:2015}. Specifically, in this work, we focus on the affine transformations. The STN consists of a localization network, a grid generator and a differentiable image sampler. The localization network is a convolutional neural network with parameter $\vtheta_{t}$, it takes as input an image $\rvx$, and regress the affine transformation parameters $\vphi$. The grid generator takes as input $\vphi$, and generates the transformed pixel coordinates as follows:

\begin{equation}
\begin{bmatrix}
u \\v
\end{bmatrix}
= \vphi
\begin{bmatrix}
\tilde{u}\\
\tilde{v}\\
1
\end{bmatrix}
=
\begin{bmatrix}
\phi_{11} & \phi_{12} & \phi_{13}\\
\phi_{21} & \phi_{22} & \phi_{23}
\end{bmatrix}
\begin{bmatrix}
\tilde{u}\\
\tilde{v}\\
1
\end{bmatrix},
\label{eqn:affine}
\end{equation}
where $(\tilde{u}, \tilde{v})$ are the normalized transformed coordinates
in the output image, $(u, v)$ are the normalized source coordinates in the input image, {\ie}, $-1 \leq \tilde{u}, \tilde{v}, u, v \leq 1$. Finally, the differentiable image sampler takes the set of sampling points from the grid generator, along with the input image $\rvx$, and produces the sampled output image $\tilde{\rvx}$. 
The bilinear interpolation is used during the sampling process.
We can denote STN by $\mathcal{T}: \rvx \rightarrow \tilde{\rvx}$ a differentiable neural network with parameter of $\vtheta_{t}$, which applies an affine transformation to the input image $\rvx$.

The goal of the adversarial geometric transformation is to find the worst-case transformations, which is equivalent to maximize following objective function:

\begin{equation}
      \label{eq:argmax_t}
      \argmax_{\vtheta_{t}} \E_{\rvx \in \mathit{D}_s \cup \mathit{D}_t} [\KL(\hat{p}(y \mid \rvx), p(y \mid \mathcal{T}({\rvx}))].
\end{equation}

The straightforward way to solve the maximization problem in \Eqref{eq:argmax_t} is to apply the gradient reverse trick, {\ie} the gradient reversal layer (GRL) in \cite{DANN:2016}, which is popular in domain adversarial training methods. The GRL has no parameters associated with it. During the forward-propagation, it acts as an identity transformation. During the back-propagation however, the GRL takes the gradient from the subsequent level and changes its sign, {\ie}, multiplies it by $-1$, before passing it to the preceding layer. Formally, the forward and back-propagation of GRL can be written as $\mathcal{R}(\rvx) = \rvx, \frac{d \mathcal{R}}{d\rvx} = - \mathbf{I}$. The loss function of the adversarial spatial transformer for domain adaptation thus can be written by:

\begin{equation}
\small
\label{eq:adv_stn_loss}
\Ls_{adv}(\vtheta_{t}) = \E_{\rvx \in \mathit{D}_s \cup \mathit{D}_t} [\KL(\hat{p}(y \mid \rvx), p(y \mid \mathcal{R}(\mathcal{T}({\rvx})))],
\end{equation}
where $\mathcal{T}$ is the spatial transformer network with parameter $\vtheta_{t}$, and $\mathcal{R}$ is the gradient reverse layer (GRL) \cite{DAN:2015}. For domain generalization, the only difference is that only $\rvx \in \mathit{D}_s$ is involved in \Eqref{eq:adv_stn_loss}. With the adversarial spatial transformer network, the final objective function for domain adaptation is written by:

\begin{equation}
      \label{eq:loss_da}
      \begin{array}{c}
      \mathcal{J}_{DA}(\vtheta_{m}, \vtheta_{t})  = \min_{\vtheta_{m}} \max_{\vtheta_{t}}(\mathcal{L}_{m} + \lambda_c \mathcal{L}_{c} + \\
      \lambda_e \mathcal{L}_{e} + \lambda_t \mathcal{L}_{adv}), \\
      \end{array}
\end{equation}
and the final objective function for domain generalization is written by:

\begin{equation}
\small
      \label{eq:loss_dg}
      \begin{array}{c}
      \mathcal{J}_{DG}(\vtheta_{m}, \vtheta_{t})  = \min_{\vtheta_{m}} \max_{\vtheta_{t}}(\mathcal{L}_{m} + \lambda_c \mathcal{L}_{c} + \lambda_t \mathcal{L}_{adv}).
      \end{array}
\end{equation}

\section{Experiments}
\label{sec:experiment}
In this section, we conduct experiments to evaluate the proposed method and compare the results with the state-of-the-art domain adaptation and generalization methods.

\begin{table*}[h!]
	\begin{center} \small
		\addtolength{\tabcolsep}{-1pt} 
		\begin{tabular}{lccl}
			\toprule
			\textbf{Methods} & \textbf{Task} & \textbf{Year} & \textbf{Description}  \\
			\midrule
			\textbf{DANN} \cite{Gani:2015}           & DA & 2015 & Domain adversarial training. \\
			\textbf{DAN} \cite{DAN:2015}             & DA & 2015 & Deep adaptation network. \\
			\textbf{ADDA} \cite{ADDA:2017}           & DA & 2017 & Adversarial discriminative domain adaptation.\\
			\textbf{JAN} \cite{JAN:2017}             & DA & 2017 & Joint adaptation network. \\
			\textbf{Dial} \cite{carlucci:2017}       & DA & 2017 &  Domain alignment layers. \\
			\textbf{DDiscovery} \cite{mancini:2018}  & DA & 2018 & Domain discovery. \\
			\textbf{CDAN} \cite{CDAN:2018}           & DA & 2018 & Conditional domain adversarial training. \\
			\textbf{MDD} \cite{mdd:2019}             & DA & 2019 & Adversarial training with margin disparity discrepancy.\\
			\textbf{Rot} \cite{rot:2019}             & DA & 2019 & Self-supervised learning by rotation prediction.\\
			\textbf{RotC} \cite{ss-da-consistency:2019} & DA & 2020 & Self-supervised learning with consistency training.\\
			\textbf{ALDA} \cite{ALDA:aaai20} & DA & 2020 & Adversarial-learned loss for domain adaptation.\\
			\textbf{MLADA} \cite{FANG2021298} 			& DA & 2021 & Multi-layer adversarial domain adaptation.\\
			\textbf{GPDA} \cite{SUN2021152} 			& DA & 2021 & Geometrical preservation and distribution alignment.\\
			
			\midrule
			\textbf{CCSA} \cite{doretto2017}         & DA \& DG  & 2017 & Embedding subspace learning. \\
			\textbf{JiGen} \cite{carlucci:2019}      & DA \& DG & 2019 & Self-supervised learning by solving jigsaw puzzle.\\
			\textbf{JigRot} \cite{jigrot:pami21}      & DA \& DG & 2021 & Self-supervised learning by combining jigsaw and rotation.\\
			
			\midrule
			
			\textbf{TF}  \cite{pacs:2017}            & DG & 2017 & Low-rank parametrized network. \\
			
			\textbf{SLRC} \cite{Ding2017DeepDG}      & DG & 2017 & Low rank constraint.\\
			\textbf{CIDDG} \cite{Li_2018_ECCV}       & DG & 2018 & Conditional invariant deep domain generalization. \\
			
			\textbf{MMD-AAE} \cite{Li_2018_CVPR}     & DG & 2018 & Adversarial auto-encoders.\\
			\textbf{D-SAM} \cite{Antonio_GCPR18}     & DG & 2018 & Domain-specific aggregation modules.   \\
			\textbf{MLDG} \cite{MLDG_AAA18}          & DG & 2018 & Meta learning approach.\\
			\textbf{MetaReg} \cite{metareg:2018}     & DG & 2018 & Meta learning approach.\\
			\textbf{MMLD} \cite{MMLD:aaai20}      & DG & 2020 & Mixture of Multiple Latent Domains.\\
			\textbf{ER} \cite{NEURIPS2020_ER}      & DG & 2020 & Domain generalization via entropy regularization.\\
			\textbf{DADG} \cite{CHEN2022418}     & DG & 2021 & Discriminative adversarial domain generalization.\\
			\textbf{WADG} \cite{ZHOU2021469}     & DG & 2021 & Wasserstein adversarial domain generalization.\\    \bottomrule
		\end{tabular}
		\caption{The compared state-of-the-art methods on domain adaptation (DA) and domain generalization (DG) task. The column \textbf{Year} shows the published year of each method.}
		\label{table:methods}
	\end{center}\vspace{-4mm}
\end{table*}

\subsection{Datasets}

Our method is evaluated on the following popular domain adaptation and generalization datasets:

\textbf{PACS} \cite{pacs:2017} is a standard dataset for DG. It contains $9991$ images collected from Sketchy, Caltech256, TU-Berlin and Google Images. It has $4$ domains (Photo, Art Paintings, Cartoon and Sketches) and each domain consists of $7$ object categories. Following \cite{carlucci:2019}, we evaluate both domain generalization and multi-source domain adaptation on this dataset.

\textbf{ImageCLEF-DA} \cite{Saenko:2010} is a benchmark dataset in the domain adaptation community for the ImageCLEF 2014 domain adaptation challenge. It consists of three domains, including Caltech-256 (C), ImageNet ILSVRC 2012 (I), and Pascal VOC 2012 (P). Each domain consist of $12$ common classes. Six domain adaptation tasks are evaluated on ImageCLEF: $\textbf{I} \rightarrow \textbf{P}$, $\textbf{P} \rightarrow \textbf{I}$, $\textbf{I} \rightarrow \textbf{C}$, $\textbf{C} \rightarrow \textbf{I}$, $\textbf{C} \rightarrow \textbf{P}$ and $\textbf{P} \rightarrow \textbf{C}$. There are $600$ images in each domain and $50$ images in each category.

\textbf{Office-Home} \cite{deephash:2017} is used for evaluating both domain adaptation and generalization. It contains 4 domains and each domain consists of images from 65 categories of everyday objects. The 4 domains are: Art (\textbf{A}), Clipart (\textbf{C}), Product (\textbf{P}) and Real-World (\textbf{R}). The Clipart domain is formed with clipart images. The Art domain consists of artistic images in the form of paintings, sketches, ornamentation \etc. The Real-world domain’s images are captured by a regular camera and the Product domain’s images have no background. The total number of images is about $15500$. 

\textbf{VLCS} \cite{Torralba:2011} is used for evaluating domain generalization. It contains images of $5$ object categories shared by $4$ separated domains: PASCAL VOC 2007, LabelMe, Caltech and Sun datasets. Different from Office-Home and PACS which are related in terms of domain types, VLCS offers different challenges because it combines object categories from Caltech with scene images of the other domains.

\textbf{VisDA}\footnote{http://ai.bu.edu/visda-2017/} is a simulation-to-real domain adaptation dataset, which has over $280K$ images across $12$ classes. The synthetic domain contains renderings of 3D models from different angles and with different lighting conditions, and the real domain contains nature images.

To investigate the robustness of the proposed model, we also evaluate on popular robustness benchmarks, including:

\textbf{CIFAR-10.1} \cite{cifar10.1} is a new test set of \textbf{CIFAR-10} with $2000$ images and the exact same classes and image dimensionality. Its creation follows the creation process of the original \textbf{CIFAR-10} paper as closely as possible. The purpose of this dataset is to investigate the distribution shifts present between the two test sets, and the effect on object recognition. 

\textbf{CIFAR-10-C} \cite{cifar10-c} is a robustness benchmark where $15$ types of corruption are algorithmically simulated to mimic real-world corruption as much as possible on copies of the \textbf{CIFAR-10} \cite{cifar10} test set.
The $15$ types of corruption are from four broad categories: noise, blur, weather and digital. Each corruption type comes in five levels of severity, with level 5 the most severe. In this work, we evaluate with the level 5 severity.

\subsection{Experimental Setting}

We implement the proposed method using the PyTorch framework on a single RTX 2080 Ti GPU with $11$ GB memory. Alexnet \cite{krizhevsky:2012}, Resnet-18 and Resnet-50 \cite{resnet:2016} architectures are used as base networks and initialized with ImageNet \cite{Deng:2009} pretrained weights. 

For training the model, we use an SGD solver with an initial learning rate of  $0.001$. We train the model for $60$ epochs and decay the learning rate to $0.0001$ after $80\%$ of the training epochs. For training baseline models, we use simple data augmentation protocols by random cropping, horizontal flipping and color jittering.

We follow the standard protocol for unsupervised domain adaptation \cite{Gani:2015} and \cite{JAN:2017} where all labeled source domain examples and all unlabeled target domain examples are used for adaptation tasks. We also follow the standard protocol for domain generalization transfer tasks as \cite{carlucci:2019} where the target domain examples are unavailable in the training phase. We set three different random seeds and run each experiment three times. The final result is the average over the three repetitions.

We compare our proposed method with state-of-the-art DA and DG methods. The descriptions of the compared methods are shown in Table~\ref{table:methods}. In the following, we use {\em{Deep All}} to denote the baseline model trained with all available source domain examples when all the introduced domain adaptive conditions are disabled. For the compared methods in Table~\ref{table:methods}, we use the results reported from the original papers if the protocol is the same.

\subsection{Experimental Results}

\subsubsection{Unsupervised domain adaptation}

The multiple source domain adaptation results on \textbf{PACS} are reported in Table~\ref{table:pacs_da}. We follow the settings in \cite{carlucci:2019} and \cite{rot:2019} and trained our model considering three domains as source datasets and the remaining one as target. {\em RotC} is an improved version {\em Rot}, which applies consistency loss with the simplest image rotation transformations \cite{ss-da-consistency:2019}. We use the open source of \cite{CDAN:2018} to produce the results of {\em CDAN} and {\em CDAN+E} and the open source of \cite{mdd:2019} to produce the results of {\em MDD}. Our proposed approach outperforms all baseline methods on all transfer tasks. The last column shows the average accuracy on the $4$ tasks. Our proposed approach outperforms state-of-the-art {\em CDAN+E} \cite{CDAN:2018} by $4.7$ percentage points and {\em MDD} by $1.4$ percentage point. 

To investigate the improvement from data augmentation, we add the same type of data augmentation as ours on {\em Deep All}, {\em DANN}, {\em CDAN}, {\em CDAN+E} and {\em MDD}, which are denoted by {\em Deep All (Aug)}, {\em DANN (Aug)}, {\em CDAN (Aug)}, {\em CDAN+E (Aug)} and {\em MDD (Aug)}. From these results, we can see that data augmentation can obtain $1.2$ to $2.6$ percentage point improvement for existing domain adaptation methods. Even with the same type of data augmentation, our proposed method still outperforms these baselines.

\begin{table}[h!]
	\begin{center} \small
		\resizebox{\linewidth}{!}{
			\begin{tabular}{@{}c@{}c@{}c@{~}c@{~}c@{~}c@{~}c}
				\toprule
				\multicolumn{2}{@{}c@{}}{\textbf{PACS-DA}}  & \textbf{art\_paint.} & \textbf{cartoon} &  \textbf{sketches} & \textbf{photo} &   \textbf{Avg.}\\ \midrule
				%
				\multirow{3}{*}{\cite{mancini:2018}}& Deep All & 74.70 & 72.40 & 60.10 & 92.90 & 75.03\\
				& Dial & 87.30 & 85.50 & 66.80 & 97.00 &  84.15 \\
				& DDiscovery  &  87.70	& 86.90	& 69.60 & 97.00		& 85.30\\
				\hline
				\multirow{2}{*}{\cite{carlucci:2019}}& Deep All & 77.85  & 74.86  & 67.74  & 95.73  & 79.05 \\
				& \hspace{-3mm}JiGen &  84.88 &	81.07 &	79.05 &	97.96 &	85.74 \\
				\hline
				\multirow{2}{*}{\cite{rot:2019}}& Deep All & 74.70 & 72.40 & 60.10 & 92.90 & 75.00 \\
				& Rot & 88.70 & 86.40 & 74.90 & 98.00 & 87.00 \\ \hline
				\multirow{2}{*}{\cite{ss-da-consistency:2019}}& Deep All & 74.70 & 72.40 & 60.10 & 92.90 & 75.00 \\
				& RotC & 90.30 & 87.40 & 75.10 & 97.90 & 87.70 \\ \hline
				
				\multirow{2}{*}{\cite{jigrot:pami21}} & Deep All & 77.83 & 74.26 & 65.81 & 95.71 & 78.40 \\
				& JigRot & 89.67 &  82.87 & 83.93 & \textbf{98.17} & 88.66 \\
				\hline
				\multirow{2}{*}{\cite{CDAN:2018}} & DANN & 82.91 & 83.83 & 69.50 & 96.29 & 83.13 \\ 
				& DANN (Aug)& 89.01 & 83.06 & 78.54 & 97.25 & 86.96 \\  \hline
				
				\multirow{2}{*}{\cite{CDAN:2018}} & CDAN & 85.70 & 88.10 & 73.10 & 97.20 & 86.00 \\
				& CDAN+E & 87.40 & 89.40 & 75.30 & 97.80 & 87.50 \\
				& CDAN (Aug)& 90.67 & 85.96 & 80.50 & 97.43 & 88.64 \\
				& CDAN+E (Aug)& 90.28 & 85.41 & 81.37 & 98.08 & 88.78 \\
				\hline
				
				\multirow{2}{*}{\cite{mdd:2019}} & MDD & 89.60 & 88.99 & \textbf{87.35} & 97.78 & 90.92 \\ 
				& MDD (Aug)& 90.28 & 86.26 & 85.72 & 97.54 & 89.95 \\  \hline
				& Deep All & 77.26  & 72.64  & 69.05  & 95.41  & 78.59 \\
				& Deep All (Aug) & 80.03  & 74.49  & 67.85  & 95.27  & 79.41 \\
				& \textbf{Ours} & \textbf{92.56} & \textbf{91.44} & 87.08 & {98.04} & \textbf{92.28} \\
				\bottomrule
		\end{tabular}}
		\caption{Multi-source domain adaptation results on PACS (Reset-18). Each column title indicates the name of the domain used as target. We use the bold font to highlight the best results.}
		\label{table:pacs_da}
	\end{center}\vspace{-8mm}
\end{table}

The results on \textbf{Office-Home} dataset are reported in Table~\ref{table:officehome_da}. On \textbf{Office-Home} dataset, we conducted $12$ transfer tasks of $4$ domains in the context of single source domain adaptation. We achieve state-of-the-art performance on $8$ out of $12$ transfer tasks.
It is noted that although \textbf{Office-Home} and \textbf{PACS} are related in terms of domain types, the number of total categories in \textbf{Office-Home} and \textbf{PACS} are $65$ and $7$ respectively. From the results, we can see that the proposed method scales when the number of categories changes from $7$ to $65$. The average accuracy achieved by our proposed method is $67.6\%$ which outperforms all the compared methods. 

\begin{table*}
	\begin{center}\small
		\addtolength{\tabcolsep}{-5pt} 
		\resizebox{\textwidth}{!}{%
			\begin{tabular}{lccccccccccccc}
				\toprule
				\textbf{Office-Home}  & 
				$\textbf{A}\hspace{-1mm}\xrightarrow{}\hspace{-1mm}\textbf{C}$ & $\textbf{A}\hspace{-1mm}\xrightarrow{}\hspace{-1mm}\textbf{P}$ & $\textbf{A}\hspace{-1mm}\xrightarrow{}\hspace{-1mm}\textbf{R}$ & $\textbf{C}\hspace{-1mm}\xrightarrow{}\hspace{-1mm}\textbf{A}$ &  $\textbf{C}\hspace{-1mm}\xrightarrow{}\hspace{-1mm}\textbf{P}$ & $\textbf{C}\hspace{-1mm}\xrightarrow{}\hspace{-1mm}\textbf{R}$ & $\textbf{P}\hspace{-1mm}\xrightarrow{}\hspace{-1mm}\textbf{A}$ & $\textbf{P}\hspace{-1mm}\xrightarrow{}\hspace{-1mm}\textbf{C}$ & $\textbf{P}\hspace{-1mm}\xrightarrow{}\hspace{-1mm}\textbf{R}$ & $\textbf{R}\hspace{-1mm}\xrightarrow{}\hspace{-1mm}\textbf{A}$ & $\textbf{R}\hspace{-1mm}\xrightarrow{}\hspace{-1mm}\textbf{C}$ & $\textbf{R}\hspace{-1mm}\xrightarrow{}\hspace{-1mm}\textbf{P}$ & \textbf{Avg.} \\ \midrule
				ResNet-50 \cite{resnet:2016} &34.9 & 50.0 & 58.0 & 37.4 & 41.9 & 46.2 & 38.5 & 31.2 & 60.4 & 53.9 & 41.2 & 59.9 & 46.1 \\
				DAN \cite{DAN:2015} & 43.6 & 57.0 & 67.9 & 45.8 & 56.5 & 60.4 & 44.0 & 43.6 & 67.7 & 63.1 & 51.5 & 74.3 & 56.3 \\
				DANN \cite{Gani:2015} & 45.6 & 59.3 & 70.1 & 47.0 & 58.5 & 60.9 & 46.1 & 43.7 & 68.5 & 63.2 & 51.8 & 76.8 & 57.6 \\
				JAN \cite{JAN:2017} & 45.9 & 61.2 & 68.9 & 50.4 & 59.7 & 61.0 & 45.8 & 43.4 & 70.3 & 63.9 & 52.4 & 76.8 & 58.3 \\
				GPDA \cite{SUN2021152} & 52.9 & \textbf{73.4} & \textbf{77.1} & 52.9 & 66.1 & 65.6 & 52.9 & 44.9 & {76.1} & 65.6 & 49.7 & 79.2 & 63.0 \\
				CDAN \cite{CDAN:2018} & 49.0 & 69.3 & 74.5 & 54.4 & 66.0 & 68.4 & 55.6 & 48.3 & 75.9 & 68.4 & 55.4 & 80.5 & 63.8 \\
				Rot \cite{rot:2019} & 50.4 & 67.8 & 74.6 & 58.7 & 66.7 & 67.4 & 55.7 & 52.4 & {77.5} & 71.0 & 59.6 & 81.2 & 65.3 \\
				CDAN+E \cite{CDAN:2018} & 50.7 & {70.6} & 76.0 & 57.6 & {70.0} & {70.0} & 57.4 & 50.9 & 77.3 & 70.9 & 56.7 & 81.6 & 65.8 \\
				ALDA \cite{ALDA:aaai20} & 53.7 & 70.1 & 76.4 & 60.2 & \textbf{72.6} & \textbf{71.5} & 56.8 & 51.9 & 77.1 & 70.2 & 56.3 & 82.1 & 66.6 \\
				RotC \cite{ss-da-consistency:2019} & 51.7 & 69.0 & 75.4 & 60.4 & 70.3 & 70.7 & 57.7 & 53.3 & \textbf{78.6} & 72.2 & 59.9 & 81.7 & 66.7 \\
				\textbf{Ours} & \textbf{55.1} & 69.0 & 74.5 & \textbf{62.5} & 66.7 & 69.8 & \textbf{62.2} & \textbf{56.0} & {77.7} & \textbf{73.5} & \textbf{61.9} & \textbf{82.2} & \textbf{67.6}\\
				\bottomrule
		\end{tabular}}
		\caption{Accuracy (\%) on Office-Home for unsupervised domain adaptation (Resnet-50). The bold font highlights the best domain adaptation results. $\textbf{A}\hspace{-1mm}\xrightarrow{}\hspace{-1mm}\textbf{C}$ indicates A (Art) is the source domain and C (Clipart) is the target domain.}
		\label{table:officehome_da}
	\end{center}
\end{table*}

The results on \textbf{ImageCLEF-DA} dataset are reported in Table~\ref{tab:image_clef_da}. As the three domains in \textbf{ImageCLEF-DA} are of equal size and balanced in each category, and are visually more similar, there is little room of improvement in this dataset. Even though, our method still outperforms the comparison methods on $4$ out of $6$ transfer tasks. 
Our method achieved $88.2\%$ average accuracy, outperforming the latest methods including {\em CDAN+E} \cite{CDAN:2018}, RotC \cite{ss-da-consistency:2019} and MLADA \cite{FANG2021298}.

\begin{table*}
	\begin{center}\small
		\setlength{\tabcolsep}{12pt}{
			\begin{tabular}{lcccccccc}
				\toprule
				\textbf{ImageCLEF-DA} & $\textbf{I}\rightarrow\textbf{P}$ & $\textbf{P}\rightarrow\textbf{I}$ & $\textbf{I}\rightarrow\textbf{C}$ & $\textbf{C}\rightarrow\textbf{I}$ &  $\textbf{C}\rightarrow\textbf{P}$ & $\textbf{P}\rightarrow\textbf{C}$ & \textbf{Avg.} \\ \midrule
				ResNet-50 \cite{resnet:2016} &74.8 &83.9 &91.5 &78.0 &65.5 &91.2 &80.7 \\
				DAN \cite{DAN:2015} & 74.5 &82.2 &92.8 &86.3 &69.2 &89.8 &82.5 \\
				Rot \cite{rot:2019} &77.9 &91.6 &95.6 &86.9 &70.5 &94.8 &84.2 \\
				DANN \cite{Gani:2015} & 75.0 &86.0 &96.2 &87.0 &74.3 &91.5 &85.0 \\
				JAN \cite{JAN:2017} & 76.8 &88.0 &94.7 &89.5 &74.2 &91.7 &85.8 \\
				CDAN \cite{CDAN:2018} & 76.7 & 90.6 & 97.0 & 90.5 & 74.5 & 93.5 & 87.1 \\
				MLADA \cite{FANG2021298}  &78.2 &91.2 &95.5 &90.8 &76.0 &92.2 &87.3 \\
				CDAN+E \cite{CDAN:2018} & 77.7 & 90.7 & \textbf{97.7} & {91.3} & 74.2 & 94.3 & 87.7 \\
				RotC \cite{ss-da-consistency:2019}  &\textbf{78.6} &92.5 &96.1 &88.9 &73.9 &\textbf{95.9} &87.7 \\
				\textbf{Ours} &{78.1} &\textbf{92.7} &96.5 &\textbf{91.6} &\textbf{74.9} &\textbf{95.9} &\textbf{88.2} \\
				
				\bottomrule
		\end{tabular}}
		\caption{Accuracy (\%) on ImageCLEF-DA for unsupervised domain adaptation (Resnet-50). The bold font highlights the best domain adaptation results. $\textbf{I}\rightarrow\textbf{P}$ indicates ImageNet ILSVRC 2012 is source domain and Pascal VOC 2012 is target domain.}
		\label{tab:image_clef_da}
	\end{center}\vspace{-8mm}
\end{table*}

The proposed method also obtains strong results on \textbf{VisDA} as reported in Table~\ref{table:visda17}. It outperforms {\em CDAN+E} by $2.6$ percentage points.

\begin{table}[htpb]
	\begin{center} \small
		\setlength{\tabcolsep}{20pt}
		\begin{tabular}{c | c}
			\toprule
			Method & Accuracy \\
			\hline
			JAN \cite{JAN:2017} & 61.6 \\
			GTA \cite{GTA:2018} & 69.5 \\
			CDAN \cite{CDAN:2018} &  66.8 \\
			CDAN+E \cite{CDAN:2018} & 70.0 \\
			Ours & \textbf{72.6} \\
			\bottomrule
		\end{tabular}
		\caption{Accuracy (\%) on VisDA (Synthetic $\rightarrow$ Real) for unsupervised domain adaptation (ResNet-50).}
		\label{table:visda17}
	\end{center}
\end{table}

It is important to understand the improvement of the consistency loss. Because {\em RotC} is the combination of simple image rotation transformation and consistency loss, without using complex data augmentation, we can better understand how much improvement is from the consistency loss. Comparing to {\em Rot}, which does not use consistency loss, we can see from the above DA experiments that {\em RotC} obtains about $0.7$ to $3.5$ percentage points improvement thanks to the consistency loss.

\subsubsection{Domain generalization}

In the context of multi-source domain generalization, we conduct $4$ transfer tasks on \textbf{PACS} dataset. We compare the performance of our proposed method against several recent domain generalization methods. We evaluated with both Alexnet and Resnet-18 and report the results in Table~\ref{table:pacs_dg_alexnet} and Table~\ref{table:pacs_dg_resnet}. From the results, we can observe that our proposed method achieves state-of-the-art domain generalization performance with both backbone architectures. With Alexnet, our method outperforms the comparison methods on all $4$ transfer tasks. The average accuracy of our method outperforms the prior best method {\em WADG} by around $1.7$ percentage points, setting a new state-of-the-art performance. With Resnet-18, the average accuracy of our method is $82.73\%$, also outperforming the existing latest ones.

As the consistency loss is not mandatory for DG, we replace consistency loss by the cross-entropy loss and run our methods. The results are denoted by \textbf{Ours w/o consis.}. We can see that the model trained with cross-entropy loss obtains similar accuracy to ours with consistency loss. However, the consistency loss is required for DA because of the unlabeled target domain samples. To keep a unified framework for both DA and DG problems, we use consistency loss for DG in this work. 

To investigate the improvement of pure data augmentation without consistency. We run {\em JiGen} and {\em Deep All} with the same type of data augmentation as ours, denoted by {\em JiGen (Aug)} and {\em Deep All (Aug)}. We can see that using pure augmentation, {\em Deep All} can obtain around $0.8$ to $1.2$ percentage point improvement, {\em JiGen} can obtain around $1.2$ to $1.6$ percentage point improvement. Even though, our proposed method still outperforms these baselines.

\begin{table}[htb]
	\begin{center} \small
		\resizebox{\linewidth}{!}{
			\begin{tabular}{@{}c@{~}c@{~}c@{~}c@{~}c@{~}c@{~}c}
				\toprule
				\multicolumn{2}{c}{\textbf{PACS-DG}}  & \textbf{art\_paint.} & \textbf{cartoon} &  \textbf{sketches} & \textbf{photo} &   \textbf{Avg.}\\ \midrule
				%
				\multirow{2}{*}{\cite{pacs:2017}}  & Deep All & 63.30 & 63.13 & 54.07 & 87.70 & 67.05\\
				& TF & 62.86 & 66.97 & 57.51  & 89.50 & 69.21\\
				\midrule
				\multirow{3}{*}{\cite{Li_2018_ECCV}} & Deep All & 57.55  & 67.04  & 58.52  & 77.98  & 65.27 \\
				& DeepC &  62.30 & 69.58  & 64.45  &  80.72 &  69.26\\
				& CIDDG &  62.70 & 69.73 & 64.45  &  78.65 &  68.88\\
				\midrule
				\multirow{2}{*}{\cite{MLDG_AAA18}} & Deep All & 64.91 & 64.28 & 53.08 & 86.67 & 67.24\\
				& MLDG & 66.23 & 66.88 & 58.96 &  88.00& 70.01\\
				\midrule
				
				\multirow{2}{*}{\cite{Antonio_GCPR18}} & Deep All  & 64.44 & {72.07} & 58.07 &  87.50 & 70.52 \\
				& D-SAM & 63.87 & 70.70 & 64.66 & 85.55 & 71.20\\
				\midrule
				
				
				\multirow{2}{*}{\cite{CHEN2022418}} & Deep All & 63.12 & 66.16 & 60.27 & 88.65 & 69.55\\
				& DADG & 66.21 & 70.28 & 62.18 &  89.76 & 72.11\\
				\midrule
				
				\multirow{2}{*}{\cite{metareg:2018}} & Deep All & 67.21 & 66.12 & 55.32 & 88.47 & 69.28\\
				& MetaReg & 69.82 & 70.35 & 59.26 &  91.07& 72.63\\
				\midrule
				
				\multirow{2}{*}{\cite{carlucci:2019}} & Deep All & 66.68 & 69.41 & 60.02 & {89.98}  & 71.52\\
				& JiGen & 67.63 & {71.71} & 65.18 & 89.00 & 73.38\\
				& JiGen (Aug)   & 71.53  & 69.50  &  68.06 &  91.08 &  75.04\\
				\midrule
				
				\multirow{2}{*}{\cite{jigrot:pami21}} & Deep All &  66.50 & 69.65 & 61.42 & 89.68 & 71.81\\
				& JigRot & 69.70 & 71.00 & 66.00 & 89.60 & 74.08\\
				\midrule
				\multirow{2}{*}{\cite{MMLD:aaai20}} & Deep All &  68.09 & 70.23 & 61.80 & 88.86 & 72.25\\
				& MMLD & 69.27 & 72.83 & 66.44 & 88.98 & 74.38 \\
				
				\midrule
				\multirow{2}{*}{\cite{NEURIPS2020_ER}} & Deep All & 68.35 &  70.14  &  90.83  &  64.98  &  73.57\\
				& ER & 71.34  &  70.29 &  89.92  &  71.15  &  75.67\\
				
				\midrule
				\multirow{2}{*}{\cite{ZHOU2021469}}& Deep All & 63.30 & 63.13 & 54.07 & 87.70 & 67.05\\ 
				& WADG & 70.21 & 72.51 & 70.32 & 89.81 & 75.71\\
				
				\midrule
				& Deep All & 68.26 & \textbf{74.52} & 63.65 & {90.78}  & 74.30\\
				& Deep All (Aug) & 73.73 & 70.09 & 65.79 & 92.22  & 75.45\\
				& \textbf{Ours w/o consis.} & 73.44 & 71.42 & \textbf{73.91} & 89.70 & 77.12\\
				& \textbf{Ours} & \textbf{74.02} & 72.23 & 72.36 & \textbf{91.16} & \textbf{77.44}\\
				\bottomrule
			\end{tabular}
		}
		\caption{Domain generalization results on PACS (Alexnet). For details about meaning of columns and use of bold fonts, see Table \ref{table:pacs_da}.}
		\label{table:pacs_dg_alexnet}
	\end{center}\vspace{-4mm}
\end{table}


\begin{table}
	\begin{center} \small
		\resizebox{\linewidth}{!}{
			\begin{tabular}{@{}c@{~}c@{~}c@{~}c@{~}c@{~}c@{~}c}
				\toprule
				\multicolumn{2}{c}{\textbf{PACS-DG}}  & \textbf{art\_paint.} & \textbf{cartoon} &  \textbf{sketches} & \textbf{photo} &   \textbf{Avg.}\\ \midrule
				\multirow{2}{*}{\cite{CHEN2022418}} & Deep All & 75.60 & {72.30} & 68.10 &  93.06 & 77.27\\
				& DADG & 79.89 & 76.25 & {70.51} & 94.86 & 80.38\\
				\midrule
				\multirow{2}{*}{\cite{Antonio_GCPR18}} & Deep All & 77.87 & {75.89} & 69.27 &  95.19 & 79.55\\
				& D-SAM & 77.33 & 72.43 & \textbf{77.83} & 95.30 & 80.72\\
				\midrule
				
				\multirow{2}{*}{\cite{carlucci:2019}} & Deep All & 77.85  & 74.86  & 67.74  & 95.73  & 79.05 \\
				& JiGen    & 79.42  & 75.25  & 71.35  &  \textbf{96.03} &  80.51\\
				& JiGen (Aug)   & 79.44  & 71.50  & 70.86  &  95.33 &  79.28\\
				\midrule
				\multirow{2}{*}{\cite{NEURIPS2020_ER}} & Deep All & 78.93  &  75.02  &  96.60 &  70.48  &  80.25\\
				& ER & 80.70 &  76.40 &  96.65 &  71.77 &  81.38\\
				\midrule
				
				\multirow{2}{*}{\cite{jigrot:pami21}} & Deep All  &  77.83 & 74.26 & 65.81 & 95.71 & 78.40\\
				& JigRot &81.07 & 73.97 & 74.67 & 95.93 & 81.41\\
				\midrule
				
				\multirow{2}{*}{\cite{metareg:2018}} & Deep All & 79.90 & 75.10 & 69.50 & 95.20 & 79.93\\
				& MetaReg & \textbf{83.70} & \textbf{77.20} & 70.30 &  95.50& 81.68\\
				\midrule
				\multirow{2}{*}{\cite{MMLD:aaai20}} & Deep All &  78.34 & 75.02 & 65.24 & 96.21 & 78.70\\ 
				& MMLD & 81.28 & 77.16 & 72.29 & 96.09 & 81.83 \\
				\midrule
				
				& Deep All & 77.26  & 72.64  & 69.05  & 95.41  & 78.59 \\
				& Deep All (Aug) & 80.03  & 74.49  & 67.85  & 95.27  & 79.41 \\
				& \textbf{Ours w/o consis.} & 81.84 & 75.05 & 77.01 & 95.07 & 82.24\\
				& \textbf{Ours}    & {82.32}  & 75.70  & 77.03  &  {95.87} &  \textbf{82.73}\\
				\bottomrule
		\end{tabular}}
		\caption{Domain generalization results on PACS (Resnet-18). For details about meaning of columns and use of bold fonts, see Table \ref{table:pacs_da}.}
		\label{table:pacs_dg_resnet}
	\end{center}\vspace{-4mm}
\end{table}

We also conduct experiments on \textbf{Office-Home} and \textbf{VLCS} dataset for multi-source domain generalization. Compare to \textbf{PACS}, these two datasets are more difficult and most of the recent works have only obtained small accuracy gain with respect to the {\em Deep All} baselines. The results on \textbf{Office-Home} and \textbf{VLCS} dataset are reported in Table~\ref{table:officehome_dg} and Table~\ref{table:vlcd_dg} respectively. Our proposed method outperforms the comparison methods on the four transfer tasks on \textbf{Office-Home} dataset. And the results tested on \textbf{VLCS} dataset show that our method achieve the best or close to the best performance on the four tasks, outperforming the recently proposed ones in average. It is noted that our baseline {\em Deep All} has relatively higher accuracy than other baselines. This is because we also add data augmentations such as random crop, horizontal flipping and color jittering when training {\em Deep All} models. In this case, it is fairer to compare with the proposed method which incorporates various data augmentation operations. 

\begin{table}
	\begin{center}\small
		\resizebox{\linewidth}{!}{
			\begin{tabular}{@{}l@{~}c@{~}c@{~}c@{~~}c@{~~}c@{~}@{~~}c@{~}}
				\toprule
				\multicolumn{2}{c}{\textbf{Office-Home-DG}}  & \textbf{Art} & \textbf{Clipart} &  \textbf{Product} & \textbf{Real-World} &  \textbf{Avg.}\\ \midrule
				
				\multirow{2}{*}{\cite{Antonio_GCPR18}} & Deep All  & 55.59 & 42.42 & 70.34 & 70.86 & 59.81 \\
				& D-SAM & {58.03} & 44.37 & 69.22 & 71.45 & 60.77\\
				\midrule
				& Deep All & 52.15 & 45.86 &  70.86 & {73.15} & 60.51\\
				\cite{carlucci:2019} & JiGen & 53.04 & {47.51} & {71.47} & 72.79 & {61.20}\\
				
				\cite{ZHOU2021469} &WADG & 55.34 & 44.82 & 72.03 & 73.55 & 61.44\\
				\cite{jigrot:pami21} &JigRot & 58.33 & 49.67 & 72.97 & 75.27 & 64.06\\	\midrule
				\multirow{2}{*}{\cite{CHEN2022418}} & Deep All &  54.31 & 41.41 & 70.31 & 73.03 & 59.77\\
				& DADG & 55.57 & 48.71 & 70.90 & 73.70 & 62.22\\
				\midrule
				
				& Deep All & 57.16 & 49.06 &  72.22 & {73.59} & 63.01\\
				& \textbf{Ours} & \textbf{59.20} & \textbf{54.67} & \textbf{73.21} & \textbf{73.93}& \textbf{65.25}\\
				\bottomrule
				
		\end{tabular}}
		\caption{Domain generalization results on Office-Home (Resnet-18). For details about meaning of columns and use of bold fonts, see Table \ref{table:pacs_da}.}
		\label{table:officehome_dg}
	\end{center}\vspace{-8mm}
\end{table}

\begin{table}
	\begin{center} \small
		\resizebox{\linewidth}{!}{
			\begin{tabular}{l@{~~~}c@{~~~}c@{~~~}c@{~~~}c@{~~~}cc}
				\toprule
				\multicolumn{2}{c}{\textbf{VLCS-DG}}  & \textbf{Caltech} & \textbf{Labelme} &  \textbf{Pascal} & \textbf{Sun} &   \textbf{Avg.}\\ \midrule
				%
				\multirow{3}{*}{\cite{Li_2018_ECCV}} & Deep All & 85.73	& 61.28	& 62.71	& 59.33	& 67.26  \\
				& DeepC & 87.47	& 62.60	& 63.97	& 61.51	& 68.89 \\
				& CIDDG & 88.83	& 63.06	& 64.38	& 62.10	& 69.59 \\
				\midrule
				\multirow{2}{*}{\cite{doretto2017}} & Deep All & 86.10 & 55.60 & 59.10 & 54.60	& 63.85 \\
				& CCSA & 92.30	& 62.10	& 67.10	& 59.10	& 70.15 \\
				\midrule
				\multirow{2}{*}{\cite{Ding2017DeepDG}} & Deep All &  86.67 & 58.20 & 59.10 & 57.86 & 65.46 \\
				& SLRC &  92.76	& 62.34	& 65.25	& 63.54	& 70.97 \\
				\midrule
				\multirow{2}{*}{\cite{pacs:2017}} & Deep All & 93.40 & 62.11 & 68.41 & 64.16 & 72.02\\
				& TF & 93.63 & {63.49} & 69.99 & 61.32 & 72.11\\
				\midrule
				\cite{Li_2018_CVPR} & MMD-AAE & 94.40&	62.60&	67.70&	{64.40} &	72.28\\ 
				\midrule
				\multirow{2}{*}{\cite{Antonio_GCPR18}} & Deep All  & 94.95  & 57.45  & 66.06  & {65.87}  &  71.08 \\
				& D-SAM & 91.75  & 56.95  & 58.59  & 60.84  & 67.03\\
				\midrule
				\multirow{2}{*}{\cite{jigrot:pami21}} & Deep All  &  96.15 & 59.05 & 70.84 & 63.92 & 72.49 \\
				& JigRot & 96.30 & 59.20 & 70.73 & 66.37 & 73.15\\
				\midrule
				
				\multirow{2}{*}{\cite{carlucci:2019}} & Deep All & {96.93} & 59.18 &  {71.96} & 62.57 & 72.66\\
				& JiGen & 96.93 & 60.90 & 70.62 & 64.30 & {73.19}\\
				\midrule
				\multirow{2}{*}{\cite{MMLD:aaai20}} & Deep All & 95.89 & 57.88 & 72.01 & 67.76 & 73.39\\
				& MMLD & 96.66 & 58.77 & 71.96 & 68.13 & 73.88\\
				\midrule
				
				\multirow{2}{*}{\cite{NEURIPS2020_ER}} & Deep All &97.15 & 58.07 & 73.11 & 68.79 & 74.28\\
				& ER & 96.92& 58.26 & \textbf{73.24} & \textbf{69.10} & 74.38\\
				\midrule
				
				\multirow{2}{*}{\cite{ZHOU2021469}} & Deep All &92.86 & 63.10 & 68.67 & 64.11 & 72.19\\
				& WADG & 96.68 & \textbf{64.26} & 71.47 & 66.62 & 74.76\\
				\midrule
				
				\multirow{2}{*}{\cite{CHEN2022418}} & Deep All &94.44 & 61.30 & 68.11 & 63.58 & 71.86\\
				& DADG & 96.80 & 63.44 & 70.77 & 66.81 & 74.76\\
				\midrule
				
				& Deep All & {97.72} & 63.03 &  {71.93} & 66.70 & 74.85 \\
				& \textbf{Ours} & \textbf{98.74} & 62.27 & {72.79} & {68.16} & \textbf{75.49} \\
				\bottomrule
				
			\end{tabular}
		}
		\caption{Domain generalization results on VLCS (Alexnet). For details about meaning of columns and use of bold fonts, see Table \ref{table:pacs_da}.}
		\label{table:vlcd_dg}
	\end{center}
\end{table}

\subsubsection{Robustness}

\begin{figure*}
    \centering
    \includegraphics[width=0.8\linewidth]{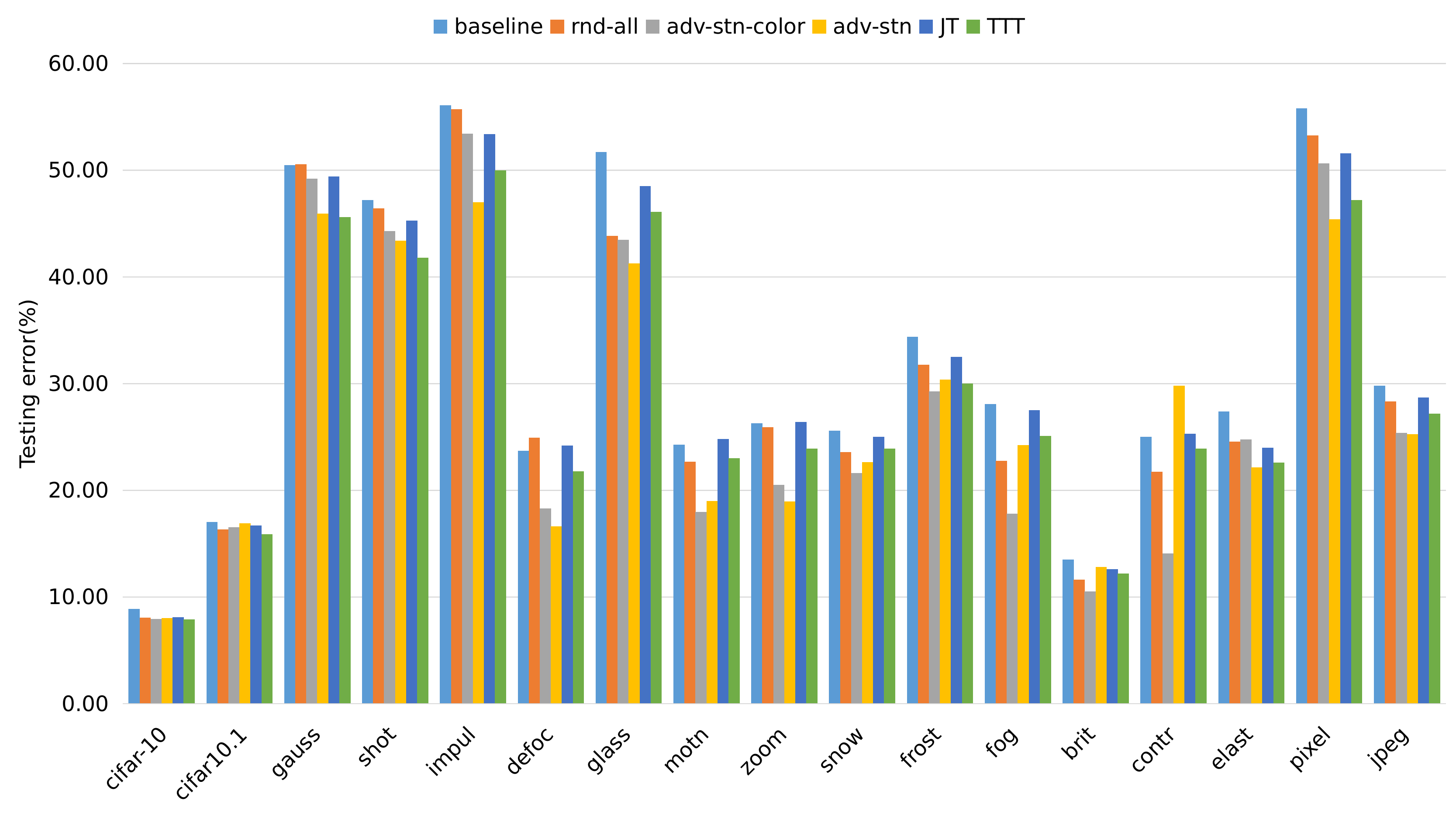}
    \caption{Test error (\%) on CIFAR-10, CIFAR10.1 and CIFAR-10-C (level 5). Best viewed in color.}
    \label{fig:cifar-10-robustness}
\end{figure*}

Apart from domain adaptation and generalization, we are also interested in the robustness of the learned model. In this part, we evaluate the proposed method on robustness benchmarks \textbf{CIFAR-10.1} and \textbf{CIFAR-10-C}. We train on the standard \textbf{CIFAR-10} training set and test on various corruption datasets, {\ie}, in a single source domain generalization setting. \figurename{ \ref{fig:cifar-10-robustness}} shows the testing error on different datasets. We evaluate different image transform strategies and also compare to recently proposed methods in \cite{Sun:2019}, {\ie}, {\em JT} and {\em TTT}. Following \cite{Sun:2019}, we use the same architecture and hyper-parameters across all experiments. 

The method denoted by {\em baseline} refers to the plain Resnet model which is equivalent to {\em Deep All} in DG setting. {\em JT} and {\em TTT} are the joint training and testing time training in \cite{Sun:2019} respectively. We denote by {\em rnd-all} the random image transformation including geometric and color-based transformations, {\em adv-stn} the proposed adversarial STN without random image transforms and {\em adv-stn-color} the adversarial STN combined with random color-based transformations. 

On the leftmost, it is the standard \textbf{CIFAR-10} testing dataset, where we can see that all the compared methods have obtained similar accuracies. On \textbf{CIFAR10.1}, the testing errors of all these methods increase simultaneously, but there is no significant gap between them. On \textbf{CIFAR-10-C} corruption data sets, the performances of these methods vary a lot. Our proposed methods show improved accuracies comparing to the {\em baseline}. {\em adv-stn-color} shows better performance than its variants and also outperforms {\em JT} and {\em TTT}. It can also be seen that {\em adv-stn} even outperforms {\em rnd-all} in most cases, although it only applies geometric transformations, which indicates the effectiveness of the proposed adversarial spatial transformations for improving robustness.

\subsection{Ablation Studies and Analysis}

We focus on the PACS DG and DA setting for an ablation analysis on the proposed method.

\subsubsection{Ablation study on image transformation strategies}

In this part, we conduct ablation studies on adversarial and random image transformations. Table~\ref{table:ablation_pacs_da} shows the ablation studies of domain adaptation on PACS with different image transformation strategies, and Table~\ref{table:ablation_pacs_dg} shows the domain generalization results. {\em rnd-color} and {\em rnd-geo} are subsets of {\em rnd-all}, where {\em rnd-color} refers to color-based transformations and {\em rnd-geo} refers to geometric transformations. Please see Table~\ref{tab:operations} for details of each subset of transformations. For DA task, when comparing individual transformation strategies, {\em rnd-color} obtained best accuracy and {\em adv-stn} outperforms {\em rnd-geo}. The combination of {\em rnd-color} + {\em adv-stn} also outperforms {\em rnd-color} + {\em rnd-geo}. However, in this experiment {\em adv-stn} does not further improve {\em rnd-color}, which might due to the limited room for improvement of the baseline method.
\begin{table*}[hpt]
	\begin{center} \small
		\setlength{\tabcolsep}{10pt}
		\begin{tabular}{ccc|cccc|c}
			\toprule
			\multicolumn{8}{c}{\textbf{PACS-DA}} \\ 
			\midrule
			\textbf{rnd-color} & \textbf{rnd-geo} & \textbf{adv-stn} & \textbf{art\_paint.} & \textbf{cartoon} &  \textbf{sketches} & \textbf{photo} &   \textbf{Avg.}\\ \hline
			$\surd$ &         &         & 93.02    	     & {91.51} & 86.62  	       & 98.00  	        & \textbf{92.29} \\
			& $\surd$ &         & 91.83     	 & 89.45   	      & 83.42  	       & 97.98  	    & 90.67 \\
			&         & $\surd$ & 91.85	         & \textbf{91.61} & 82.45 	& {97.92} 	    & 90.96 \\
			\hline
			$\surd$ & $\surd$ &         & \textbf{93.10} & 91.01   	      & 86.33  	       & \textbf{98.14}   	     & 92.15   \\
			$\surd$ &         & $\surd$ & {92.56}        & {91.44}        & \textbf{87.08} & {98.04}         & \textbf{92.28} \\
			\bottomrule
			
		\end{tabular} 
		\caption{Ablation studies of domain adaptation on PACS. The first three columns indicate the types of image transformations applied. Each column title in the middle indicates the name of the domain used as target. We use the bold font to highlight the best results.}
		\label{table:ablation_pacs_da}
	\end{center}
\end{table*}

\begin{table*}[h]
	\begin{center} \small
		\setlength{\tabcolsep}{10pt}{ 
			\begin{tabular}{ccc|cccc|c}
				\toprule
				\multicolumn{8}{c}{\textbf{PACS-DG}} \\ 
				\midrule
				\textbf{rnd-color} & \textbf{rnd-geo} & \textbf{adv-stn} & \textbf{art\_paint.} & \textbf{cartoon} &  \textbf{sketches} & \textbf{photo} &   \textbf{Avg.}\\ \hline
				$\surd$ &         &         & 71.40 	     & 72.43 	      & \textbf{71.44} & 90.20 	        & 76.37 \\ 
				& $\surd$ &         & 71.08 	     & 72.40 	      & 66.98 	       & 91.36 	        & 75.46 \\
				&         & $\surd$ & 70.92	         & \textbf{73.46} & 69.66 	       & 90.78 	        & 76.21 \\
				\hline
				$\surd$ & $\surd$ &         & 73.05 	     & 72.15 	      & 69.08 	       & \textbf{91.64} & 76.48 \\
				$\surd$ &         & $\surd$ & \textbf{74.02} & {72.23} & \textbf{72.36} & {91.16} & \textbf{77.44}\\
				\bottomrule
				
		\end{tabular} }
		\caption{Ablation studies of domain generalization on PACS. The first three columns indicate the types of image transformations applied. Each column title in the middle indicates the name of the domain used as target. We use the bold font to highlight the best results.}
		\label{table:ablation_pacs_dg}
	\end{center}
\end{table*}

In the DG experiment, we get similar conclusion that {\em adv-stn} outperforms {\em rnd-geo}. As the baseline accuracy of DG is far from saturated than DA, we can see that {\em adv-stn} further improves {\em rnd-color} and the combination of  {\em rnd-color} + {\em adv-stn} obtained the best accuracy.

\subsubsection{Ablation study on hyperparameters setting}

\begin{figure*}
	\centering
	\begin{tabular}{c@{~~~}c}
	\includegraphics[width=0.45\linewidth,trim=2cm 1cm 2cm 2cm,clip]{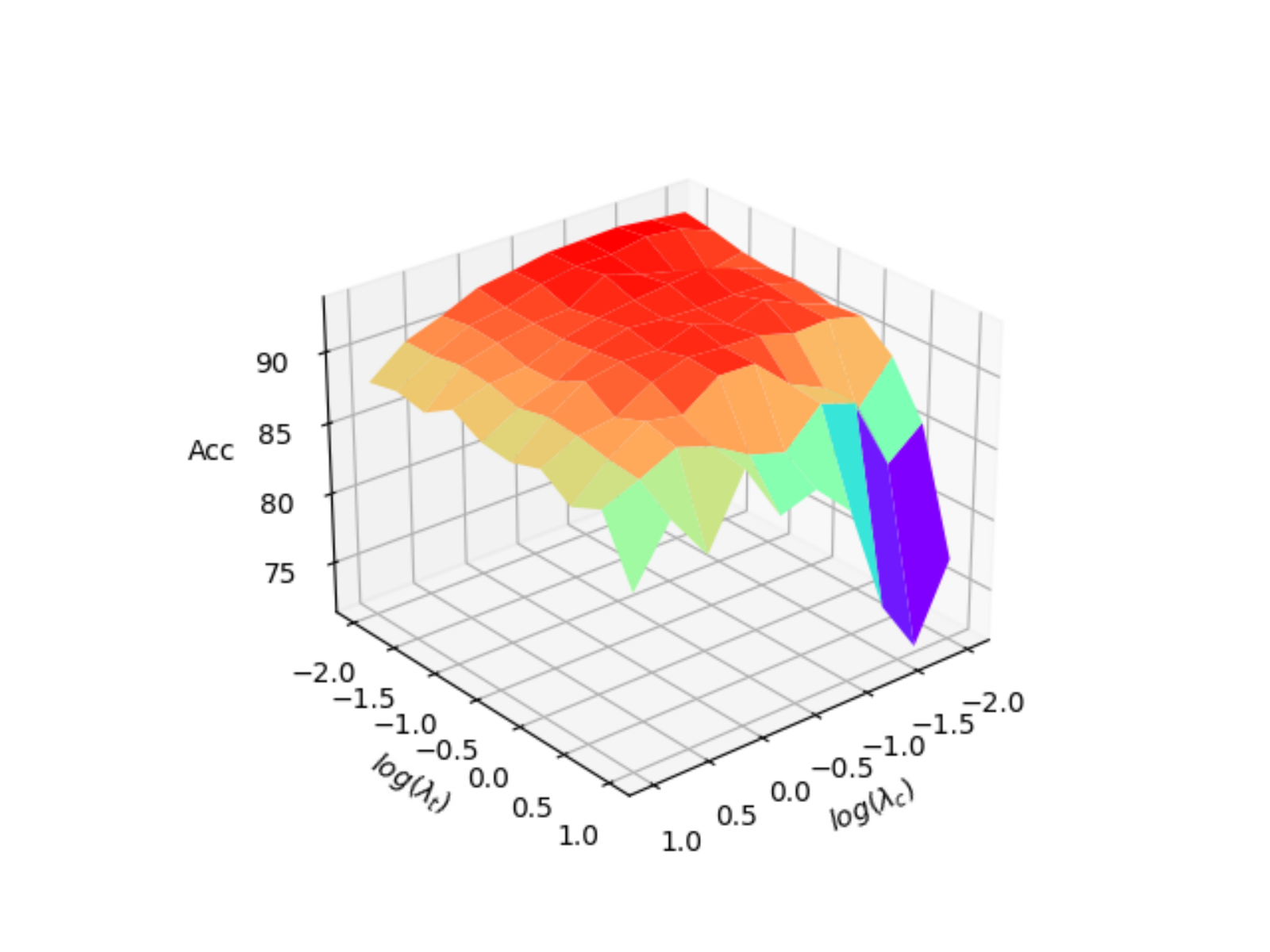} &	\includegraphics[width=0.45\linewidth,trim=2cm 1cm 2cm 2cm,clip]{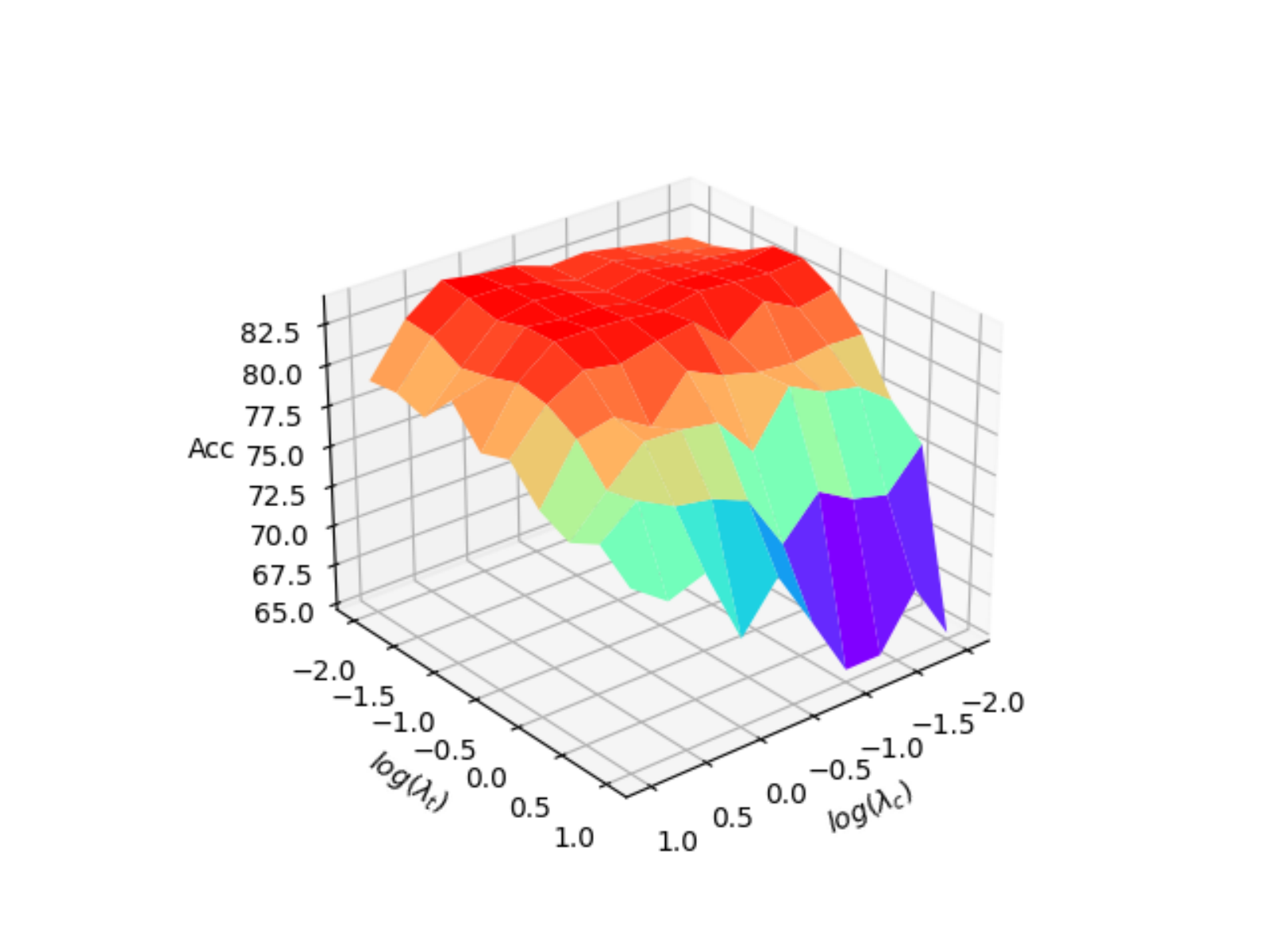} \\
	\vspace{-1em}
	(a)& (b)
	\end{tabular}
	\caption{Accuracies with different hyperparameter settings on PACS $\textbf{photo}, \textbf{cartoon}, \textbf{sketch} \rightarrow{} \textbf{art\_painting}$ task (Resnet-18). (a) Multi-source Domain adaptation with fixed $\lambda_e$. (b) Domain generalization. Best viewed in color.}
	\label{fig:ablation_hyper}
\end{figure*}

In this part, we conduct ablation studies on hyperparameters setting.
The final objective functions of our proposed method for domain adaptation (\ref{eq:loss_da}) and domain generalization (\ref{eq:loss_dg})  are weighted summation of several items, with the weighting factors as the hyperparameters. Since the conditional entropy minimization loss is widely used domain adaptation, we fixed the weight $\lambda_e=0.1$  and conducted grid test with different settings of $\lambda_c$ and $\lambda_t$, which are shared by domain generalization. We tested ten different values which are logarithmically spaced between $10^{-2}$ and $10$ for $\lambda_c$ and $\lambda_t$. For each setting, we run with three different random seeds and calculate the mean accuracy. The results of multi-source domain adaptation and domain generalization which take \textbf{photo}, \textbf{cartoon} and \textbf{sketch} as the source domains and \textbf{art\_painting} as the target domain are reported in \figurename{   \ref{fig:ablation_hyper}}. Resnet-18 is used as the base network.

From the figures, we can see that when both $\lambda_c$ and $\lambda_t$ are  not too large, the accuracies are relatively stable, validating the insensitiveness of our proposed method to hyperparameters. However, when $\lambda_c$ and $\lambda_t$ grow too large, the performance decreases, especially with a small $\lambda_c$ and a large $\lambda_t$. The reason maybe that overwhelmingly large weights for consistency loss and adversarial spatial transformer loss over the main classification loss make the learned feature less discriminative for the main classification task, therefore resulting in lower accuracy. And too large $\lambda_t$ may lead to excessive emphasis on the extreme geometric distortions, which may be harmful to the general cross domain performance.

\subsubsection{Visualization of learned deep features}

\begin{figure*}[htb]
	\centering
	\begin{tabular}{c@{~~~}c@{~~~}c@{~~~}c}
		\includegraphics[height=2.5cm]{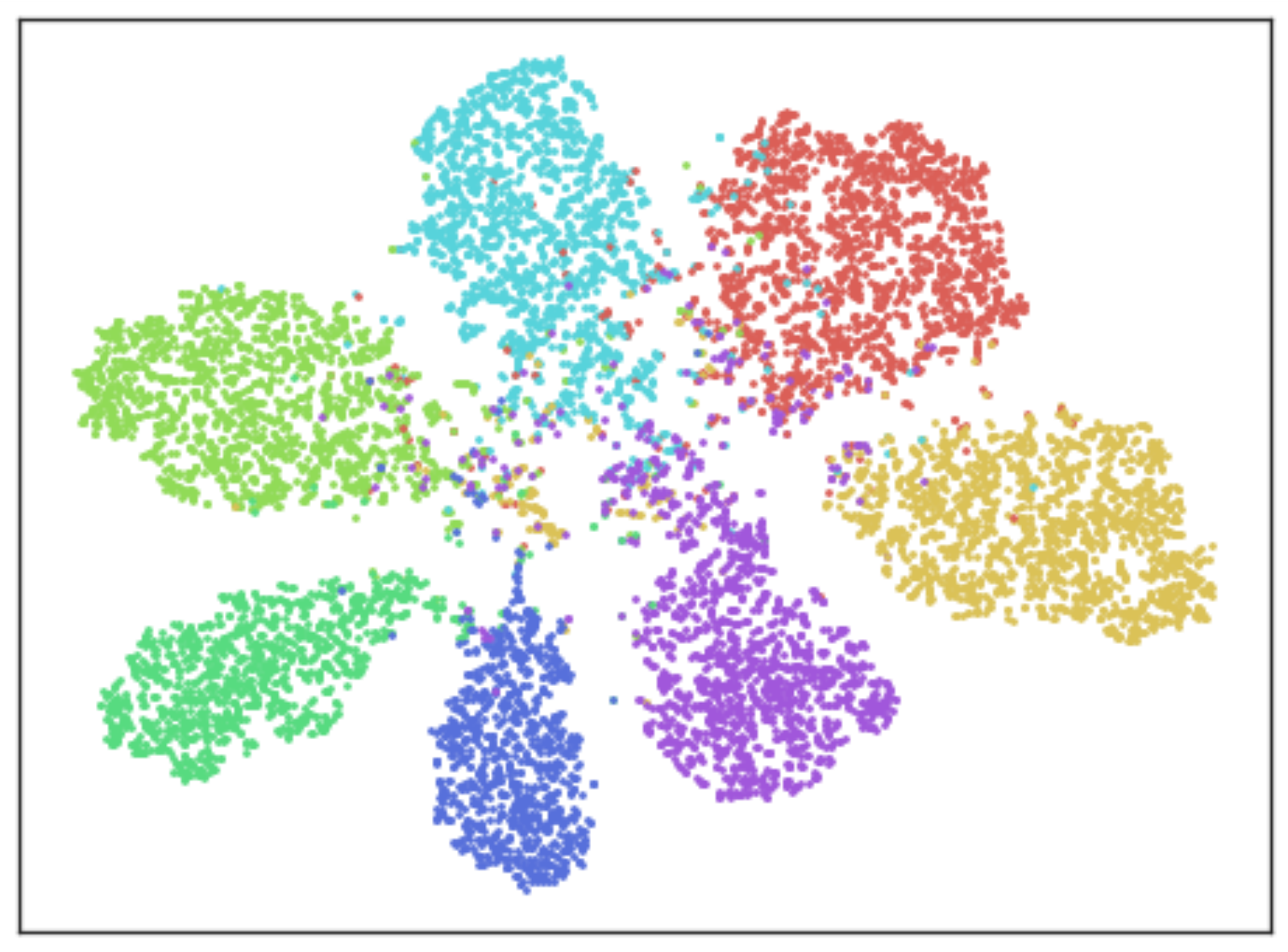}   &  \includegraphics[height=2.5cm]{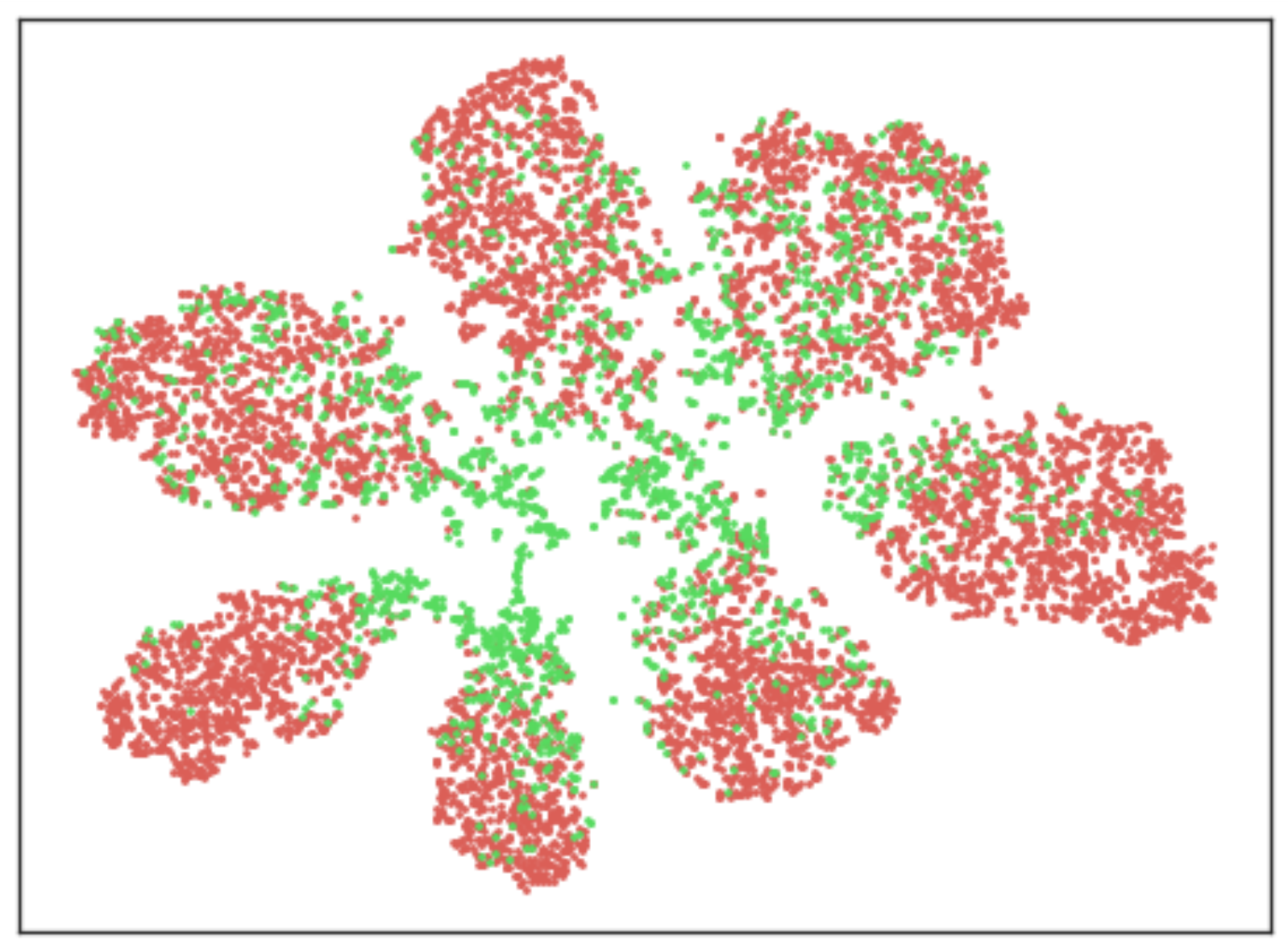} &
		\includegraphics[height=2.5cm]{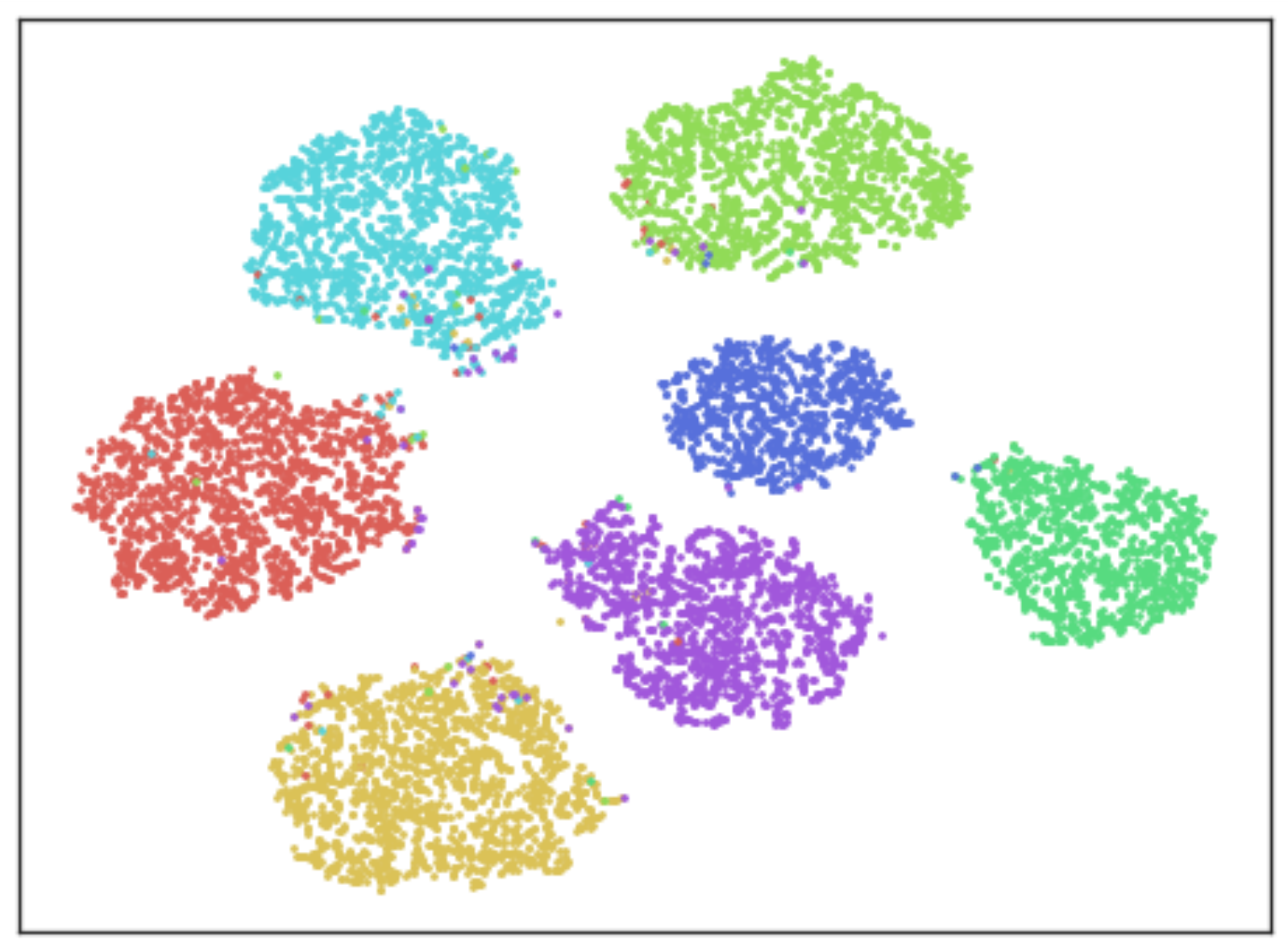}  & \includegraphics[height=2.5cm]{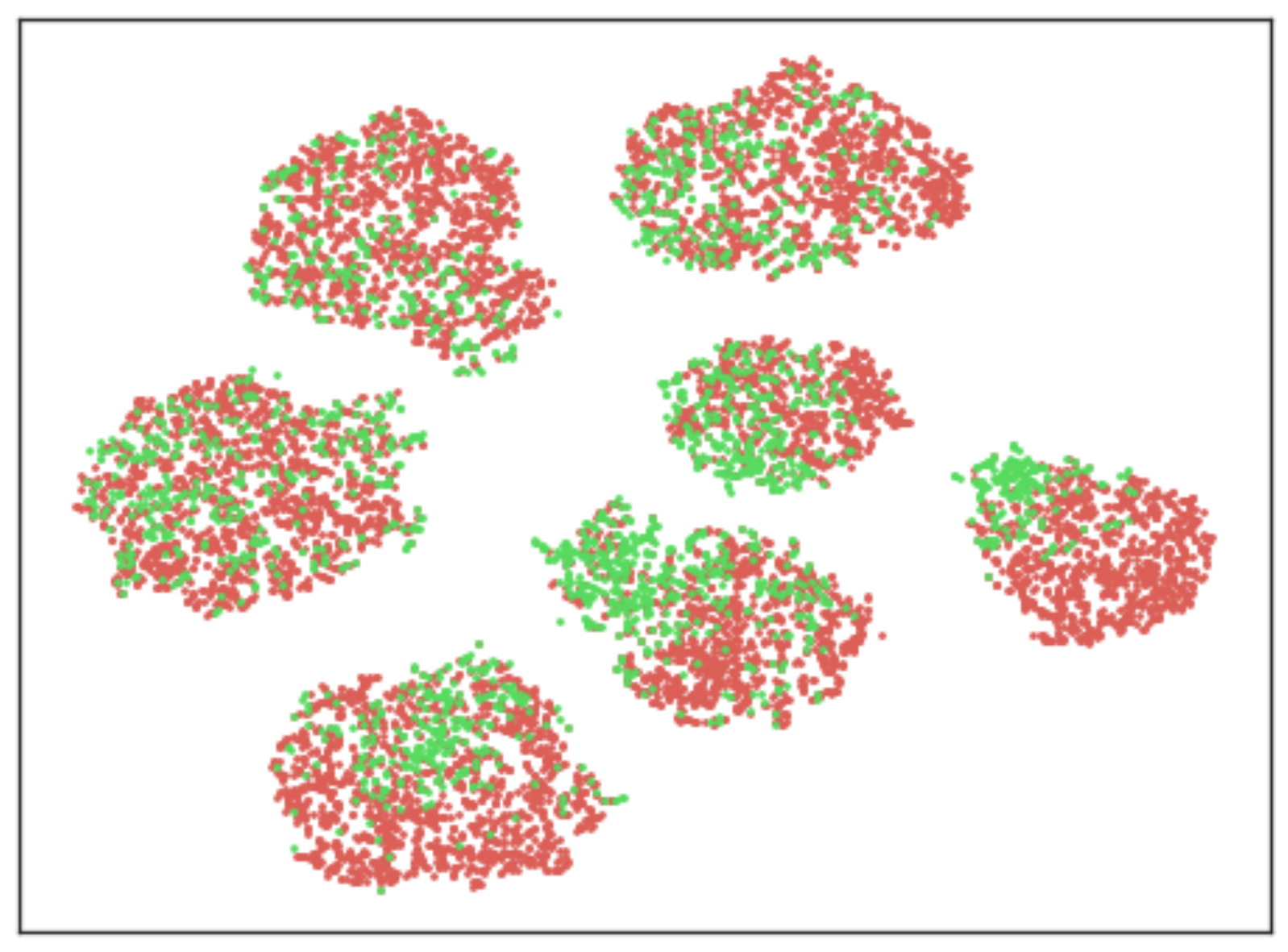}\\
		(a) & (b) & (c) & (d) \\
	\end{tabular}
	\caption{The t-SNE visualization on the PACS DA setting: (a) class visualization of {\em Deep All}, (b) domain visualization of {\em Deep All}, (c) class visualization of {\em Ours}, (d) domain visualization of {\em Ours}.}
	\label{fig:tsne_pacs_da}
\end{figure*}

\begin{figure*}[htb]
	\centering
	\begin{tabular}{c@{~~~}c@{~~~}c@{~~~}c}
		\includegraphics[height=2.5cm]{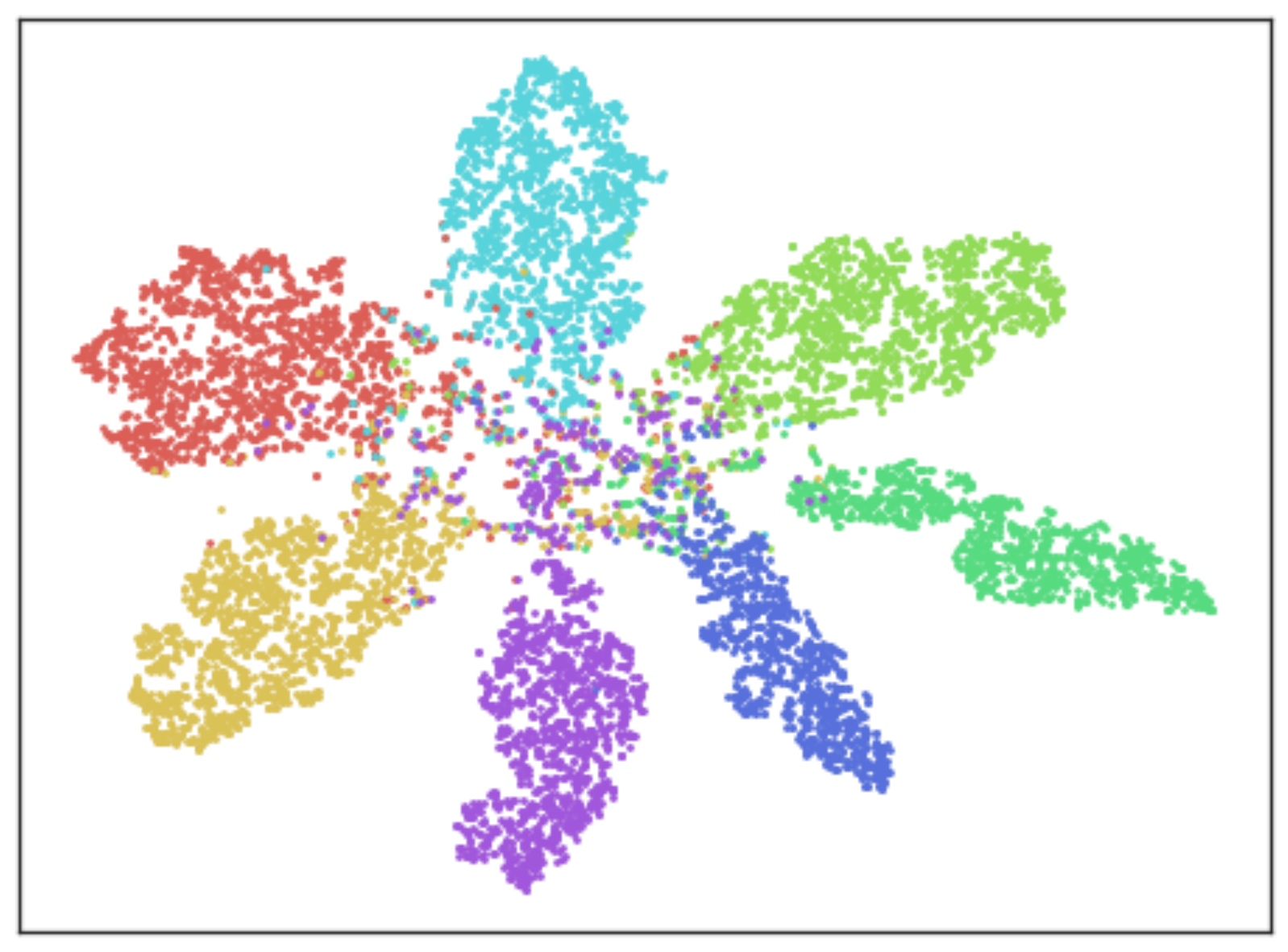}   &  \includegraphics[height=2.5cm]{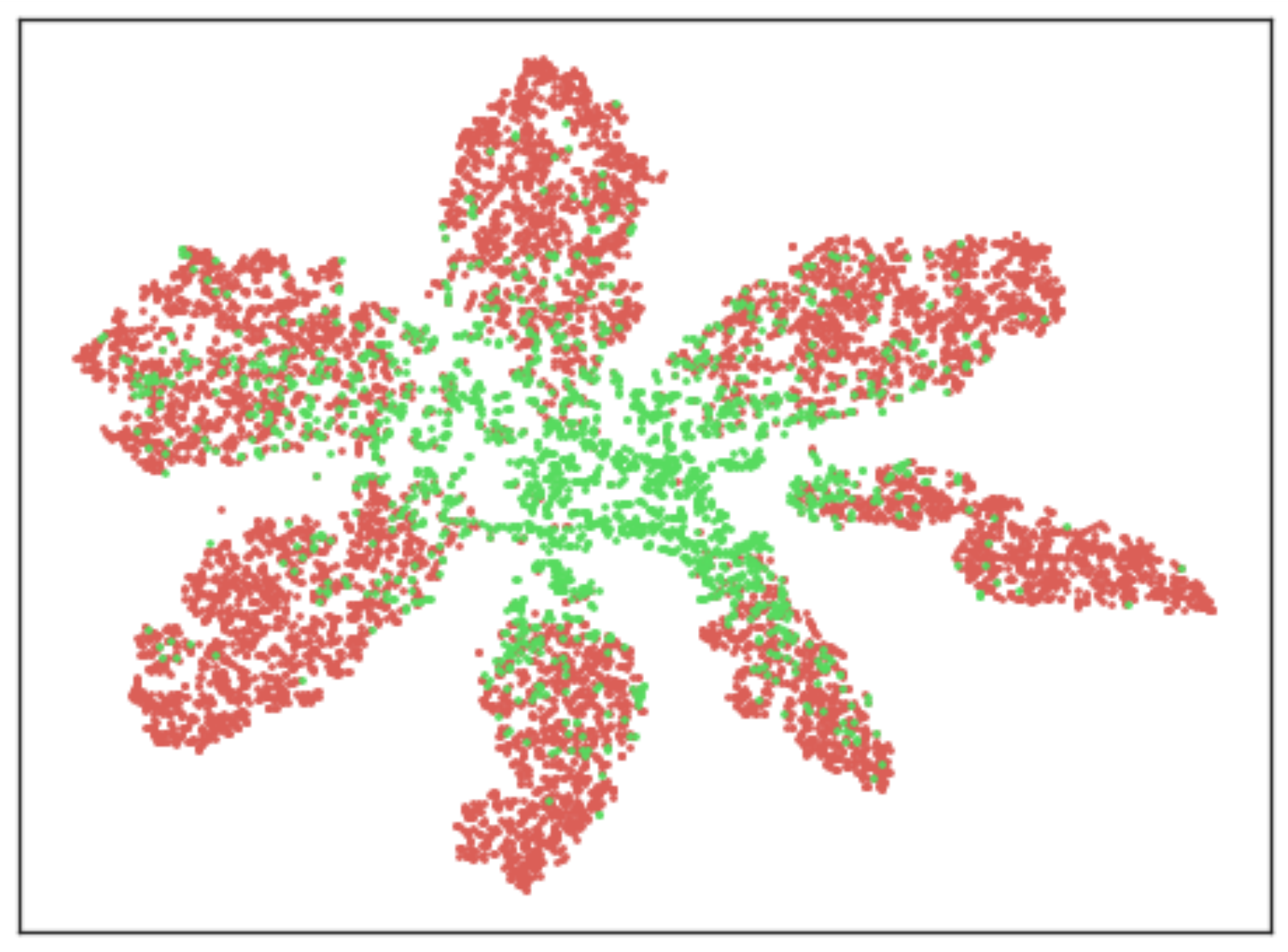} &
		\includegraphics[height=2.5cm]{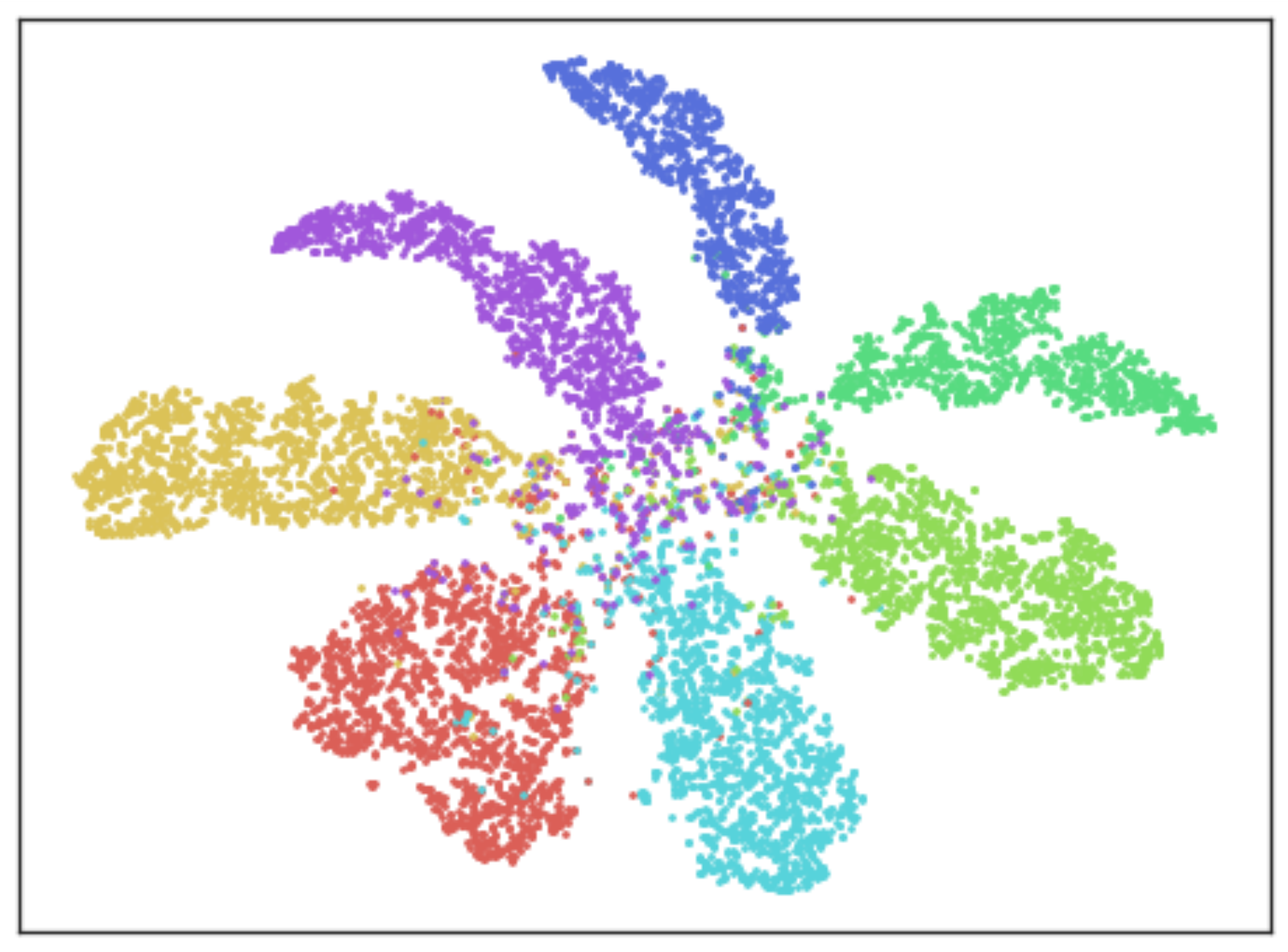}  & \includegraphics[height=2.5cm]{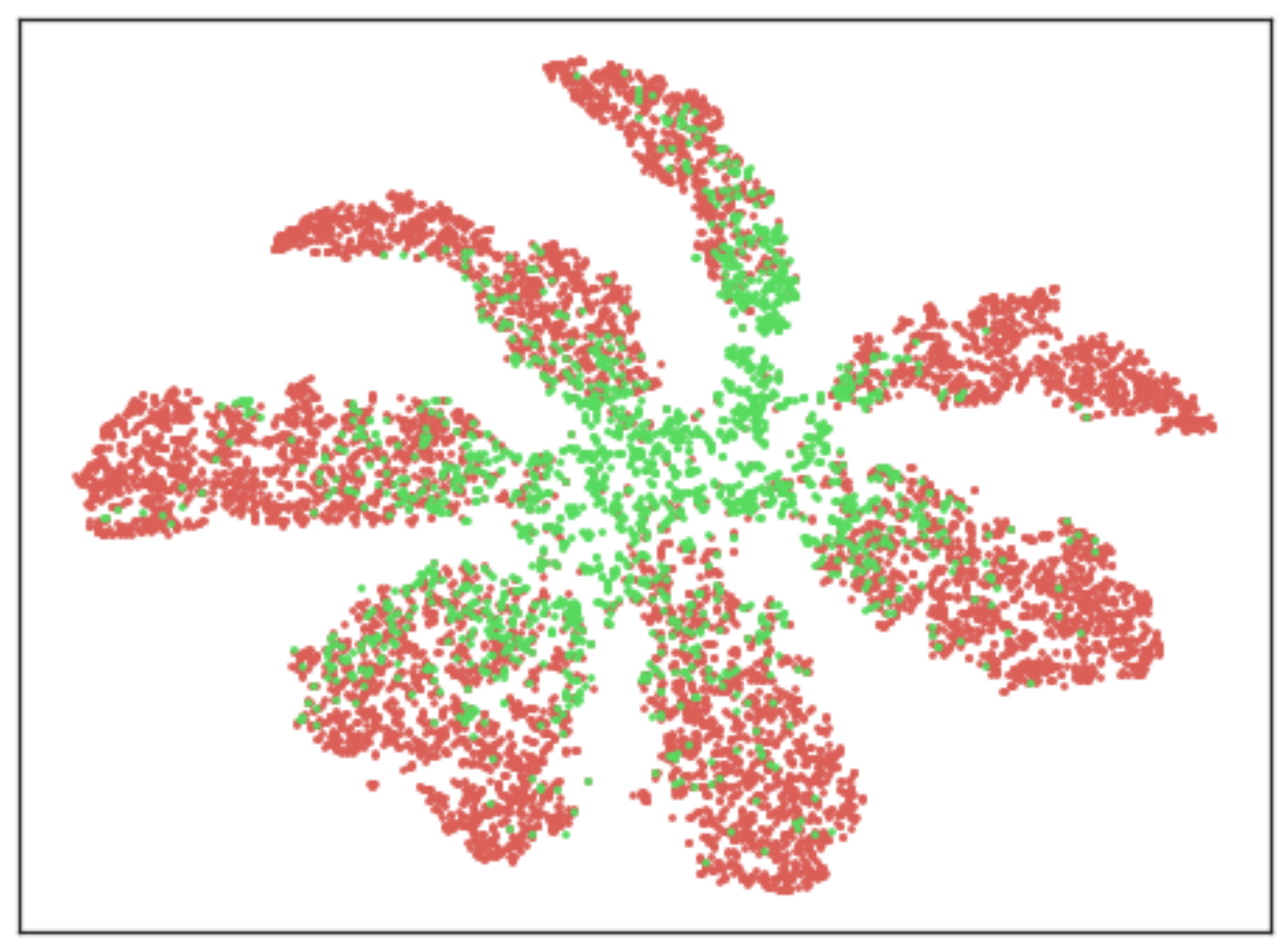}\\
		(a) & (b) & (c) & (d) \\
	\end{tabular}
	\caption{The t-SNE visualization on the PACS DG setting: (a) class visualization of {\em Deep All}, (b) domain visualization of {\em Deep All}, (c) class visualization of {\em Ours}, (d) domain visualization of {\em Ours}.}
	\label{fig:tsne_pacs_dg}
\end{figure*}

To better understand the learned domain invariant feature representations, we use t-SNE \cite{Maaten:2008} to conduct an embedding visualization. We conduct experiments on transfer task of $\textbf{photo}, \textbf{cartoon}, \textbf{sketch} \rightarrow{} \textbf{art\_painting}$ with both DA and DG settings and visualize the feature embeddings. \figurename{\ref{fig:tsne_pacs_da}} shows the visualization on PACS DA setting, and \figurename{\ref{fig:tsne_pacs_dg}} shows the visualization on PACS DG setting. On both figures, we visualize category alignment as well as domain alignment. We also compare to the baseline {\em Deep All} which does not apply any adaptation. 

From the visualization of the embeddings, we can see that the clusters created by our model not only separates the categories but also mixes the domains. The visualization from DG model suggests that our proposed method is able to learn feature representations generalizable to unseen domains. It also implies that the proposed method can effectively learn domain invariant representation with unlabeled target domain examples.

\subsubsection{Visualization of adversarial examples}

\begin{figure*}
    \centering
    \includegraphics[width=\linewidth]{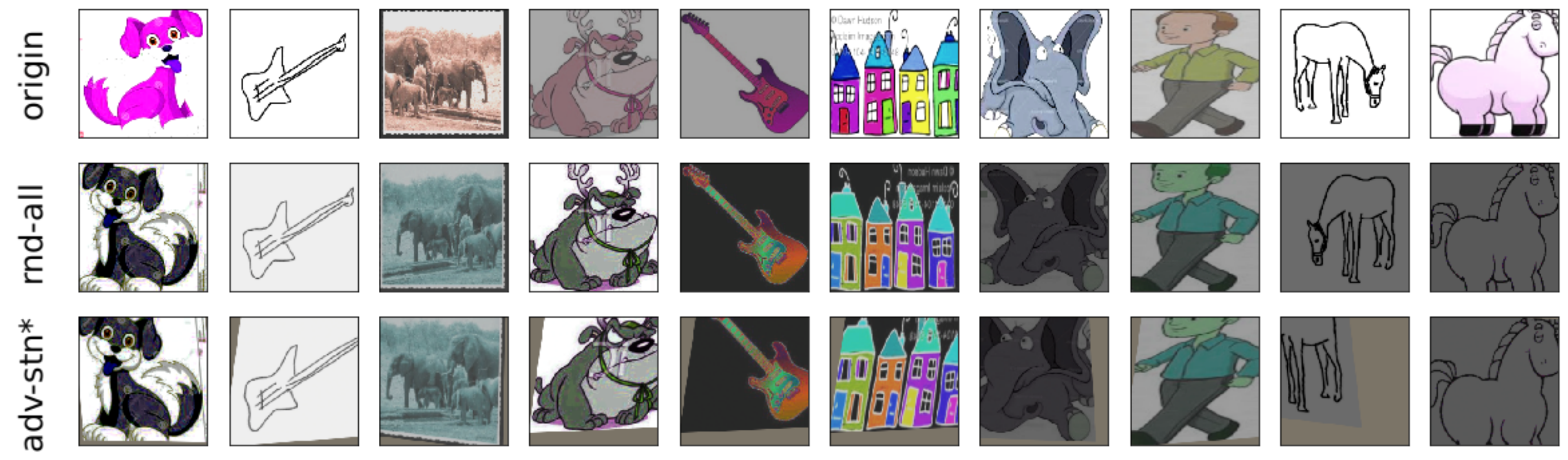}
    \caption{Visualization of the transformed images during the PACS DG training. First row: original image with random horizontal flipping and jittering; second row: original image with the proposed random augmentation; third row: the original image with the proposed adversarial spatial transformation combined with random color-based transformations.}
    \label{fig:examples_pacs_dg}
\end{figure*}

To visually examine what are the adversarial spatial transformations learned by {\em adv-stn}, we plot the transformed examples during training in \figurename{ \ref{fig:examples_pacs_dg}}. In the first row, we show the original images with simple random horizontal flipping and jittering augmentation. The second and third rows show images with {\em rnd-all} and {\em adv-stn-color} respectively. From the figure, we can see that the proposed adversarial STN does find more difficult image transformations than the random augmentation. Training with these adversarial examples greatly improves the generalization ability and robustness of the model.

\section{Conclusion}
\label{sec:conclusion}
In this work, we proposed a unified framework for addressing both domain adaptation and generalization problems. Our domain adaptation and generalization methods are built upon random image transformation and consistency training. This simple strategy can obtain promising DA and DG performance on multiple benchmarks. To further improve its performance, we proposed a novel adversarial spatial transformer networks which can find the worst-case image transformation to improve the generalizability and robustness of the model. Experimental results on multiple object recognition DA and DG benchmarks verified the effectiveness of the proposed methods.

\section*{Acknowledgments}
This work is supported by the National Natural Science Foundation of China under Grant No. 61803380 and 61790565.

\bibliographystyle{hindawi_bib_style}
\bibliography{mybibfile} 

%

\end{document}